\documentclass[10pt,twocolumn,letterpaper]{article}

\usepackage[pagenumbers]{cvpr} 

\usepackage[accsupp]{axessibility} 
\definecolor{cvprblue}{rgb}{0.21,0.49,0.74}
\usepackage[pagebackref,breaklinks,colorlinks,allcolors=cvprblue]{hyperref}
\usepackage{multirow,graphicx}
\usepackage{mathtools}
\def\paperID{12225} 
\def\confName{CVPR}
\def\confYear{2025}

\title{SplatFlow: Self-Supervised Dynamic Gaussian Splatting in Neural Motion Flow Field for Autonomous Driving}

\author{
Su Sun\thanks{Equally contributed as co-first author.}~~$^{1}$, Cheng Zhao\footnotemark[1]~~$^{2}$, Zhuoyang Sun$^{1}$, Yingjie Victor Chen$^{1}$, Mei Chen$^{2}$ \\
$^{1}$Purdue University, $^{2}$Microsoft\\
}

\begin{document}
\maketitle
\begin{abstract}
Most existing Dynamic Gaussian Splatting methods for complex dynamic urban scenarios rely on accurate object-level supervision from expensive manual labeling, limiting their scalability in real-world applications.
In this paper, we introduce SplatFlow, a Self-Supervised Dynamic Gaussian Splatting within Neural Motion Flow Fields (NMFF) to learn 4D space-time representations without requiring tracked 3D bounding boxes, enabling accurate dynamic scene reconstruction and novel view RGB/depth/flow synthesis.
SplatFlow designs a unified framework to seamlessly integrate time-dependent 4D Gaussian representation within NMFF, where NMFF is a set of implicit functions to model temporal motions of both LiDAR points and Gaussians as continuous motion flow fields.
Leveraging NMFF, SplatFlow effectively decomposes static background and dynamic objects, representing them with 3D and 4D Gaussian primitives, respectively.
NMFF also models the correspondences of each 4D Gaussian across time, which aggregates temporal features to enhance cross-view consistency of dynamic components.
SplatFlow further improves dynamic object identification by distilling features from 2D foundation models into 4D space-time representation.
Comprehensive evaluations conducted on the Waymo and KITTI Datasets validate SplatFlow's state-of-the-art~(SOTA) performance for both image reconstruction and novel view synthesis in dynamic urban scenarios. 
\vspace{-6mm}
\end{abstract}

\section{Introduction}
\label{sec:intro}
\begin{figure}[!t]
    \centering
    \includegraphics[width=1.0\columnwidth]{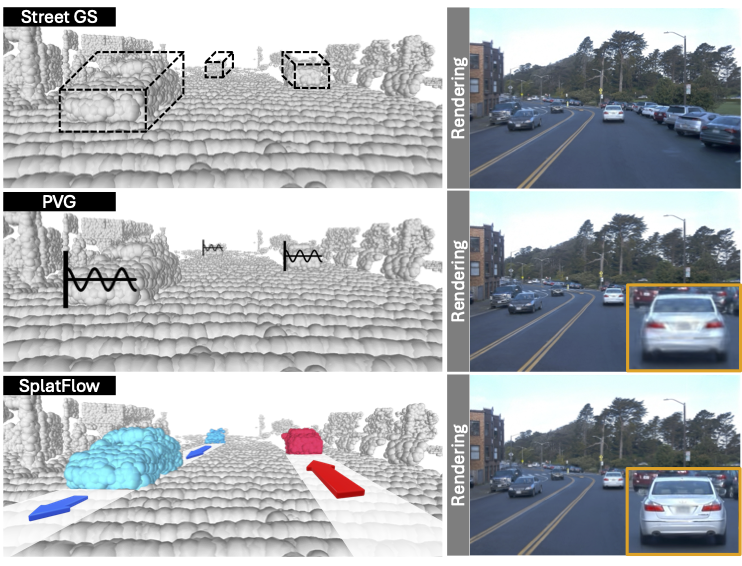}
    \vspace{-8mm}
    \caption{Top: Street GS~\cite{yan2024street};
    Middle: PVG~\cite{chen2023periodic};
    Bottom: Our SplatFlow. SplatFlow eliminates the need for 3D Bboxes required by Street GS, and enhances rendering quality compared to PVG.
    }
  \label{fig:intro}
  \vspace{-7mm}
\end{figure}

As autonomous driving systems increasingly shift toward end-to-end models, there is a growing need for scalable simulation environments without domain gaps where these systems can undergo closed-loop evaluations. 
A promising approach is real-world closed-loop evaluation, which demands controllable sensor inputs, thereby driving the development of advanced scene reconstruction techniques. 
In this context, Neural Radiance Fields (NeRFs)~\cite{mildenhall2021nerf} and 3D Gaussian Splatting (3DGS)~\cite{kerbl20233d} have proven to be effective in creating high-quality 3D scene reconstructions with excellent visual and geometric accuracy.
However, accurately and comprehensively reconstructing dynamic driving scenes remains a major challenge, given the complexity of real-world scenarios without dynamic object annotations.

Recent approaches advance NeRF-based representations by incorporating object detection and tracking to model dynamic elements, enabling photo-realistic view generation of dynamic urban street environments. 
Approaches~\cite{ost2021neural, wu2023mars, kundu2022panoptic, ziyang2023snerf, tonderski2024neurad, fischer2024multi} create a scene graph, where both dynamic objects and the static background are represented as nodes, reconstructed within their canonical frames.  
Self-supervised methods~\cite{turki2023suds,yang2023emernerf} model dynamic driving scenes by combining static neural field and time-dependent neural field to handle the static background and moving foreground objects separately.
The SUDS~\cite{turki2023suds} uses optical flow to ease the strict requirement for object labeling, while EmerNerf~\cite{yang2023emernerf} learns temporal attributes and features in a self-supervision manner to avoid reliance on optical flow.
However, these NeRF style methods face efficiency challenges in both training and inference, which becomes a major bottleneck for large-scale scene reconstruction and rendering.

While NeRF has proven effective for driving scenes, 3DGS offers a promising alternative due to faster training and rendering speed with more explicit representations. 
However, the original 3DGS faces notable challenges in modeling dynamic urban environments due to its limited representation capabilities in the temporal dimension.
To address this, DrivingGaussian~\cite{zhou2024drivinggaussian} introduces composite dynamic Gaussian graphs to handle multiple moving objects and incremental static Gaussians for background representation. 
StreetGaussian~\cite{yan2024street} optimizes the tracked bounding boxes of dynamic Gaussians along with 4D spherical harmonics to capture changing vehicle appearances. 
Both methods, however, require accurate object-level supervision, such as 3D object boxes and trackers, to decompose static and dynamic elements, limiting their scalability in real-world applications. 
In response, PVG~\cite{chen2023periodic} proposes a Periodic Vibration Gaussian mechanism to model dynamic driving scenes without relying on manually labeled 3D bounding boxes.
However, the scene decomposition in PVG relies on time-dependent Gaussian attributes optimized by rendering losses, but ignores motion cues within the input point cloud used for Gaussian initialization, leading to suboptimal performance in challenging scenarios with rapid movements.
To facilitate accurate scene reconstruction and real-time rendering without expensive annotations, we propose SplatFlow, a self-supervised Dynamic Gaussian Splatting method in Neural Motion Flow Fields~(NMFF) for dynamic urban scenarios. 
In contrast to existing methods, as illustrated in Fig.~\ref{fig:intro}, SplatFlow decomposes dynamic objects and the static background in a self-supervised way without requiring expensive 3D bounding box annotations. 
The key idea of SplatFlow is to seamlessly integrate 4D Gaussian representations within NMFF in a unified framework. 
The NMFF is a set of implicit functions to model temporal motions of both LiDAR points and Gaussians as continuous motion flow fields.
Levaraging NMFF, SplatFlow not only enables the decomposition of dynamic and static elements from 3D LiDAR points, but also enables the status conversion of each 4D Gaussian across time.
During Gaussian splatting, we represent dynamic objects using aggregated 4D Gaussians at various viewpoints and timestamps, and the static background using 3D Gaussians, respectively.
In SplatFlow, NMFF's capabilities are enhanced through three components: 1) learning 3D motion priors by pretraining on 3D LiDAR data, 2) optimizing temporal status transitions of 4D Gaussians on image data, and 3) distilling knowledge from 2D foundation models.

The novel features of SplatFlow are summarized as:
\begin{itemize}
    \item SplatFlow introduces a unified framework that seamlessly integrates time-varying 4D Gaussian representations into NMFF, enabling self-supervised dynamic scene reconstruction and rendering.
    \item NMFF enables scene decomposition, modeling static elements with 3D Gaussians and dynamic elements with 4D Gaussians.
    \item NMFF models correspondences of each 4D Gaussian over time, aggregating temporal features to enhance the cross-view consistency of dynamic components. 
    \item SplatFlow enhances the dynamic object identification by uplifting 2D features distilled from foundation models to 4D space-time through optical flow rendering.
\end{itemize}
Comprehensive experiments on the Waymo~\cite{sun2020scalability} and KITTI~\cite{geiger2012} benchmarks demonstrate that SplatFlow outperforms the state-of-the-art~(SOTA) methods in both image reconstruction and novel view synthesis for dynamic urban scenes. 
Notably, SplatFlow achieves this without relying on tracked 3D bounding boxes of dynamic objects, allowing the proposed model to learn from extensive, in-the-wild data sources.

\section{Related work}
\label{sec:related work}
\begin{figure*}[!t]
    \centering
    \includegraphics[width=1.0\textwidth]{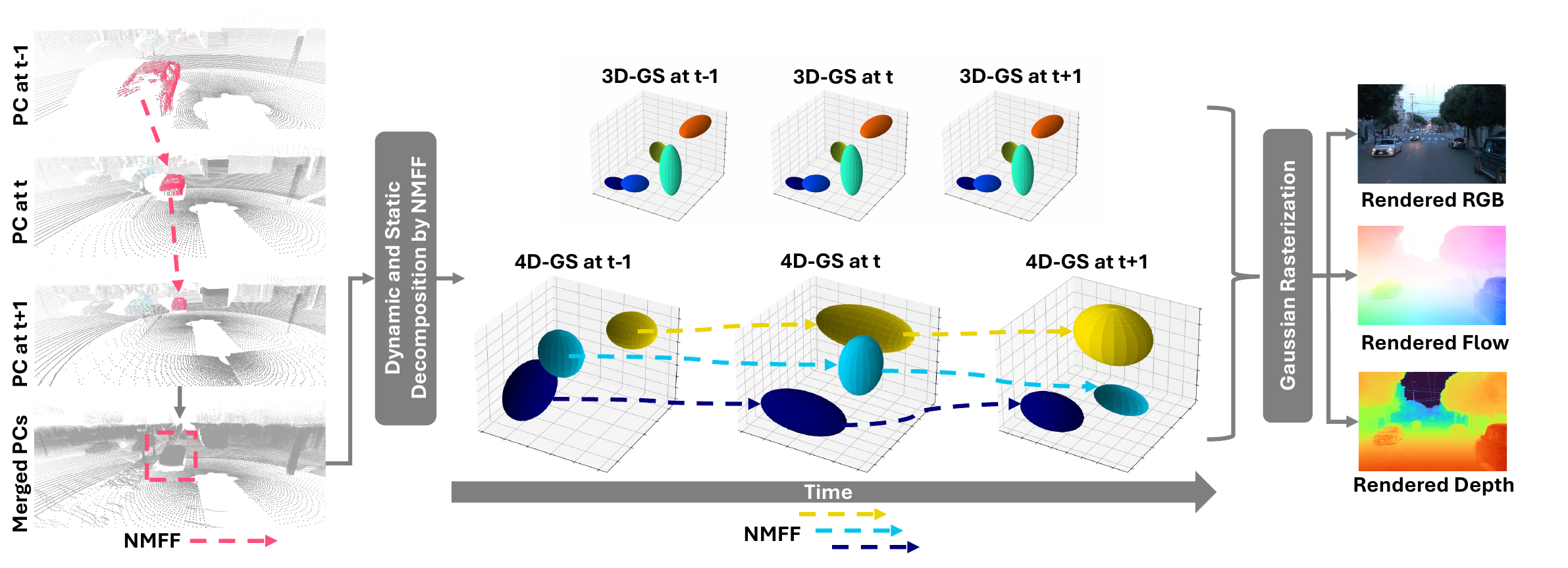}
    \vspace{-8mm}
    \caption{The pipeline of SplatFlow.}
  \label{fig:network}
  \vspace{-6mm}
\end{figure*}

Real-world simulation is essential for generating data to support end-to-end autonomous driving solutions. 
However, current simulation engines like AirSim~\cite{shah2018airsim} and CARLA~\cite{dosovitskiy2017carla} encounter difficulties due to the expensive manual work needed to create virtual environments and the domain gap between real and simulated data. 
The rapid progress in Novel View Synthesis (NVS) technologies, such as NeRF~\cite{mildenhall2021nerf} and 3DGS~\cite{kerbl20233d}, and following work ~\cite{mueller2022instant, zhang2022nerfusion, Sun_2024_CVPR, 10655755} enables 3D reconstruction and photorealistic image generation, significantly improving real-world simulation for autonomous driving.
For 3D reconstruction and neural rendering in autonomous driving scenes, sensor fusion solution combining surrounding cameras and LiDAR is commonly used in existing work~\cite{rematas2022urban, lu2023urban, ost2022neural, guo2023streetsurf, zhao2024tclc}. 
Urban Radiance Field~\cite{rematas2022urban} improves NeRF training by incorporating 3D data from LiDAR to enhance 3D geometry learning. 
DNMP~\cite{lu2023urban} employs a pre-trained deformable mesh primitive to represent the 3D scene, enhancing neural rendering quality. 
NPLF~\cite{ost2022neural} uses explicit 3D reconstructions from LiDAR data to efficiently model the radiance field.
TCLC-GS~\cite{zhao2024tclc} design a hybrid explicit and implicit 3D representation derived from LiDAR-camera data, to enrich the properties of 3D Gaussians for splatting.
Though these methods show impressive results for static urban scene rendering, they struggle with dynamic objects common in driving scenarios.

To model dynamic urban scenes, recent approaches~\cite{kundu2022panoptic, ost2021neural, wu2023mars, turki2023suds, yang2023emernerf} decompose the entire scene into static background and dynamic objects, learning their representations separately. 
PNF~\cite{kundu2022panoptic} uses monocular 3D bounding box predictions to isolate dynamic objects and further jointly optimizes their poses during the reconstruction process. 
NSG~\cite{ost2021neural} uses neural graphs to represent entire scenes, decomposing dynamic multi-object environments.
MARS~\cite{wu2023mars} employs distinct sub-networks to model backgrounds and dynamic objects, creating an instance-aware simulation framework. 
These methods typically rely on manually annotated or predicted 3D bounding boxes. 
In order to avoid the need of 3D bounding boxes, SUDS~\cite{turki2023suds} introduces a scalable hash table to represent large-scale dynamic urban scenes, using an off-the-shelf 2D optical flow estimator to track dynamic objects. 
EmerNerf~\cite{yang2023emernerf} addresses this challenge by learning scene flow to associate points across time in the 4D neural radiance field, allowing the separation of static and dynamic objects without 3D bounding boxes. 
However, these methods, which rely on implicit representations, still suffer from inefficiency in both reconstruction and rendering.
Recently, 3DGS~\cite{kerbl20233d} introduced a novel explicit 3D scene representation, combining high-quality volume rendering with a fast speed. 
However, 3DGS is designed for static scenes and fails when modeling dynamic moving objects. 
To address this, DrivingGaussian~\cite{zhou2024drivinggaussian} and StreetGaussian~\cite{yan2024street} partition the driving scene into static and dynamic components by separate sets of Gaussians based on the 3D bounding boxes of vehicles. 
DrivingGaussian~\cite{zhou2024drivinggaussian} uses a composite dynamic Gaussian graph to manage multiple moving objects while incrementally employing incremental static 3D Gaussians to represent the static background.
StreetGaussian~\cite{yan2024street} models foreground vehicles with optimizable poses and a 4D spherical harmonics appearance model, and background with 3D Gaussians separately.
HUGS~\cite{zhou2024hugs} also adopts 3D object boxes to identify dynamic elements, enabling optimization of geometry, appearance, semantics, and motion attributes for dynamic Gaussians.
However, all these methods rely on expensive annotated or predicted 3D bounding boxes to learn the time-dependent representations of dynamic objects. 
Most recent method PVG~\cite{chen2023periodic} introduces periodic vibration based Gaussian attributes, optimized through self-supervision without the need of 3D bounding boxes. 
Each Gaussian models dynamic changes over time through optimizable attributes including vibration directions, life span, and life peak.
\section{Methodology}
\label{sec:methodology}
\begin{figure*}[!t]
    \centering
    \includegraphics[width=1\textwidth]{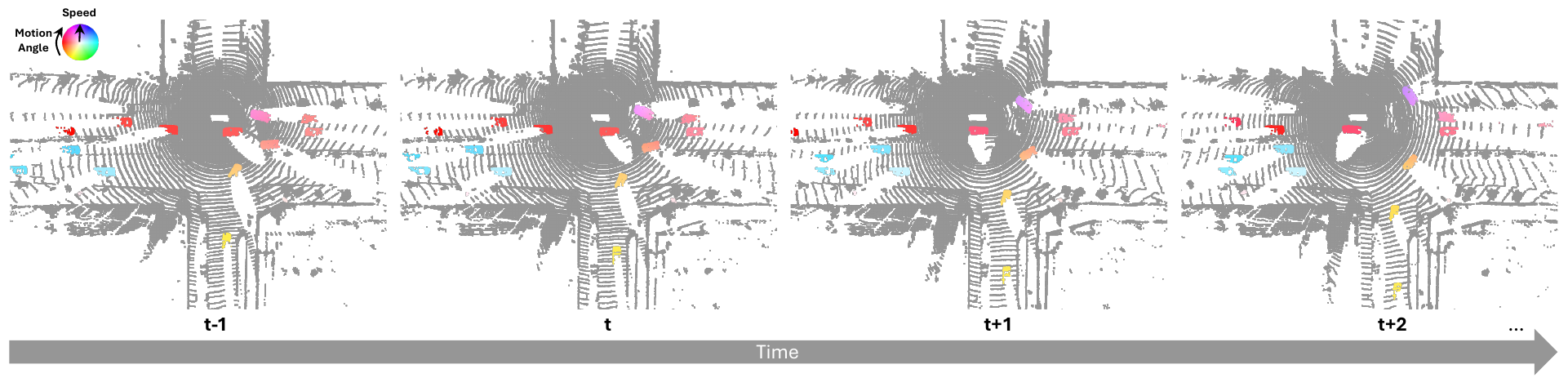}
    \vspace{-8mm}
    \caption{Visualization of 3D LiDAR points within NMFF on Waymo dataset.}
  \label{fig:LiDAR motion flow}
  \vspace{-6mm}
\end{figure*}

\subsection{Overview}
By collecting images and point clouds with timestamps from surrounding cameras and LiDAR, we aim to learn a space-time 4D Gaussian representation of a dynamic scene without any human annotations, enabling fast and high-quality novel view rendering. 
As the pipeline of SplatFlow shown in Fig.~\ref{fig:network},
NMFF models the temporal motions of both 3D points and Gaussians as continuous motion flow fields. 
It serves two key roles in SplatFlow: 1) decomposing static and dynamic elements from 3D LiDAR points; 2) modeling the temporal correspondences of each 4D Gaussian across time.
We first decompose the time-series LiDAR points into static and dynamic points by NMFF, which are then separately merged to initialize the static and dynamic Gaussians.
Dynamic objects are represented by aggregated 4D Gaussians associated with NMFF over time, while the static background is modeled using 3D Gaussians. 
NMFF learns the correspondence of each 4D Gaussian in conjunction with optimizing 4D Gaussians' attributes along time.
We also distill optical flow knowledge from a 2D foundation model into 4D space-time representation. 
Finally, images, depths and optical flows are rendered from novel viewpoints at different timesteps in dynamic driving scenes.

\subsection{Problem Definition:}
In a dynamic urban scenario, we collect time-sequential data from a vehicle equipped with surrounding cameras and a LiDAR sensor.
A set of surrounding images $\mathcal{I}$ is captured by multiple surrounding cameras with corresponding intrinsic matrices $I$ and extrinsic matrices $E$. 
Meanwhile, a set of 3D points $\mathcal{P}$ is obtained from LiDAR, along with corresponding extrinsic matrices $E'$. 
We denote the synchronized and calibrated multi-sensor data sequence, including timestamps $t$, as $\{ \mathcal{I}_i, \mathcal{P}_i, E_i, I_i, E'_i, t_i| i=1, 2, 3, \dots, n\}$, where $\mathcal{P}_i = \{ x_i, y_i, z_i \}$ and $n$ is the number of frames. 
The vehicle trajectory is either provided or estimated using multi-sensor-based odometry. 
Our model $\mathcal{SF}$ aims to perform accurate 3D reconstruction and synthesize novel viewpoints for any given timestamp $t$ and camera pose $[E_t, I_t]$ by rendering $\hat{\mathcal{I}} = \mathcal{SF}(E_t, I_t, t)$.

\subsection{4D Gaussian Representation}
Given images with associated camera poses, 3DGS optimizes a set of anisotropic 3D Gaussians through differentiable rasterization to represent a static 3D scene. 
We extend 3DGS to 4DGS of spatial-temporal representations for dynamic objects in an urban scene. 
Each 4D Gaussian primitives $\mathcal{G}(t)$ is represented by time-varying attributes: 3D center $\mu(t) = [x(t), y(t), z(t)]^T$ and covariance matrix $\Sigma(t)$ along with time-invariant attributes: opacity $\sigma$ and color $c$. 
The density of a Gaussian at point $(x, t)$ is defined as,
\vspace{-2mm}
\begin{equation}
\alpha(t) = \sigma \cdot exp ( -\frac{1}{2}(x-\mu(t))^T \Sigma(t)^{-1} (x-\mu(t))).
\label{eq:1}
\vspace{-2mm}
\end{equation}
The covariance matrix $\Sigma(t) = R(t)SS^TR(t)^T$ is composed of a scaling matrix $S = diag(s_x, s_y, s_z)$ and rotation matrix $R(t) = (q_x(t), q_y(t), q_z(t), q_w(t))$, constrained to be a positive semi-definite matrix during optimization.

To render an image from a specific viewpoint at timestamp $t$, we first transform all the 4D Gaussians from other times to the target time $t$, according to the learned correspondence from NMFF. 
Subsequently, the aggregated 4D Gaussians are splatted onto the image plane, resulting in a collection of 2D Gaussians. 
The 3D covariance matrix $\Sigma$ is projected to a 2D covariance matrix $\Sigma'$ by,
\vspace{-1mm}
\begin{equation}
\Sigma'(t) = JE \Sigma(t) E^TJ^T,
\label{eq:2}
\vspace{-1mm}
\end{equation}
where $E$ refers to the world-to-camera matrix and $J$ refers to Jacobian of the perspective transformation.
By sorting the Gaussians according to their depth within the camera space, we use $\alpha$-blending to estimate the color $C$ and depth $D$ at each pixel $p$ as,
\vspace{-2mm}
\begin{equation}
C = \sum_{i=1}^{N}  c_i \alpha_i \prod_{j=1}^{i-1}(1 - \alpha_j),
D = \sum_{i=1}^{N}  z_i \alpha_i \prod_{j=1}^{i-1}(1 - \alpha_j).
\label{eq:3}
\vspace{-2mm}
\end{equation}
Here $c_i$ represents the color of Gaussian $\mathcal{G}_i$, which is computed using spherical harmonic coefficients. 
$z_i$ represents the distance from the image plane to the $i$-th Gaussian center.
$\alpha_i$ denotes the density computed from the 2D projection of the $i$-th Gaussian and its learned opacity.

\subsection{Neural Motion Flow Field}
To establish the correspondence of dynamic objects, we design NMFF to model 3D motions as a continuous motion flow field, enabling the temporal transition of both 3D points and Gaussians.
The NMFF $\Phi$ contains a temporal sequence of motion flow field, defined as $\Phi = \{\phi_{t_i:t_{i+1}} \}, i = 0,1,2,...,n-1$, which predicts 3D motion flow of an arbitrary query point between two consecutive frames as,
\vspace{-2mm}
\begin{equation}
\phi_{t_1:t_2}(x_{t_1}, y_{t_1}, z_{t_1}) = \Delta x_{t_1:t_2},  \Delta y_{t_1:t_2},  \Delta z_{t_1:t_2}, \Delta R_{t_1:t_2}
\label{eq:4}
\end{equation}
where $\Delta R_{t_1:t_2}$ denotes motion angle between two adjacent timestamps.
Each field $\phi$ is eight ReLU-MLP stacks.

\textbf{Point Cloud within NMFF:}
Although NMFF can be learned during Gaussian Splatting optimization, we observed that the rendering loss alone is insufficient to effectively constrain the motion representation of dynamic scene components.
Therefore, we leverage the rich geometric structure inherent in temporally consecutive 3D LiDAR points to derive a robust NMFF prior, as shown in Fig.~\ref{fig:LiDAR motion flow}.

We first pre-train $\Phi$ on a sequence of 3D LiDAR points by forward and backward 3D geometry consistency, enabling NMFF to learn the 3D motion of each query 3D point over time. 
Given the source point clouds $\mathcal{P}$ and target point clouds $\mathcal{Q}$ from different timestamps, we minimize the point distance between the source and target point clouds by a bidirectional Chamfer Distance~($CD$),
\vspace{-2mm}
\begin{equation}
d_{\mathrm{CD}}(\mathcal{P}, \mathcal{Q}) 
= \sum_{p \in \mathcal{P}} \min_{q \in \mathcal{Q}} \|p - q\|^2
+ \sum_{q \in \mathcal{Q}} \min_{p \in \mathcal{P}} \|q - p\|^2.
\label{eq:5}
\vspace{-2mm}
\end{equation}
The bidirectional $CD$ is a point-based distance that establishes correspondences in both directions—source to target and target to source—by searching for the nearest neighbors in each point cloud.

We compensate for ego-car motion in LiDAR points to let NMFF learn only 3D motion flow during pre-training. 
As a result, the 3D motion flow predicted by NMFF excludes the ego-car motion.
We generate a 3D dynamic mask to separate static and dynamic points by applying a threshold on each point’s 3D motion flow. 
We then merge static point clouds using ego-motion alone, and dynamic point clouds using both ego-motion and motion flow. 
For the static background, we initialize the static 3D Gaussians using the merged 3D static points.
For dynamic objects, we initialize 4D Gaussians by aggregating points from different timestamps into a common reference frame (the mid-timestamp reference frame), which increases density and completeness by integrating observations from various viewpoints over time.
\textbf{4D Gaussian within NMFF:}
The pre-trained $\Phi$ is subsequently integrated with 4D Gaussian representations of dynamic objects, enabling joint optimization during Gaussian splatting.
Given consecutive time $t_1$ and $t_2$ where $t_1 < t_2$ with respective states $\{\mathcal{G}_i(t_1)\}$ and $\{\mathcal{G}_i(t_2)\}$, these states are connected by NMFF for each Gaussian as,
\vspace{-3mm}
\begin{equation}
\mathcal{G}_i(t_1) = \{ \mu(t_1), R(t_1), S, \alpha, c \} ,
\label{eq:5}
\vspace{-3mm}
\end{equation}
\vspace{-3mm}
\begin{equation}
\Delta \mu_{t_1:t_2}, ~ \Delta R_{t_1:t_2} = \phi_{t_1:t_2}(\mu(t_1)) ,
\label{eq:6}
\vspace{-3mm}
\end{equation}
\vspace{-4mm}
\begin{equation}
\hat{\mathcal{G}_i}(t_2) = \{ \mu(t_1) + \Delta \mu_{t_1:t_2}, R(t_1) \cdot \Delta R_{t_1:t_2}, S, \alpha, c \}.
\label{eq:7}
\vspace{-2mm}
\end{equation}

We employ NMFF $\Phi$ correspondence to warp the center of all aggregated 4D Gaussians across time to the desired timestamp, as $\{ \mathcal{G}_i(t) \} = \{\mathcal{G}(t), \hat{\mathcal{G}}(t_1), \hat{\mathcal{G}}(t_2), ..., \hat{\mathcal{G}}(t_n) \}$.
For the transition of Gaussians across multiple timestamps, we apply the corresponding NMFF through step-by-step propogation. 
In practice, we propogate the aggregated 4D Gaussian at the mid-timestamp to the target timestamp.
Here NMFF learns a motion transition of the 4D Gaussian among training frames in a self-supervised manner, promoting a multi-view consistent representation.
For static background, we represent it by static 3D Gaussians $\mathcal{G}_{static}$ following the strategy~\cite{chen2023periodic}.
For sky area, we adopt a separate environmental map with static sky Gaussians $\mathcal{G}_{sky}$ following \cite{chen2023periodic, yang2023emernerf}.
Given a specific viewpoint $[E_t, I_t, t]$, the image $\hat{\mathcal{I}}$ and depth $\hat{\mathcal{D}}$ at time $t$ are obtained by the differentiable rendering using a set of Gaussians,
\vspace{-1mm}
\begin{equation}
\hat{\mathcal{I}}, \hat{\mathcal{D}} = Render( \{\mathcal{G}_i(t)\}, \mathcal{G}_{static}, \mathcal{G}_{sky} | E_t, I_t, t).
\label{eq:8}
\end{equation}

\subsection{Optical Flow Distillation}
The 3D motion flow provided by NMFF enables 2D optical flow rendering.  
Given two consecutive time $t_1$ and $t_2$, the optical flow $f_{t_1:t_2}$ of each Gaussian is computed by projecting 3D center $\mu$ to the image plane using the camera's intrinsic matrix $I$ and extrinsic matrix $E$,
\vspace{-2mm}
\begin{equation}
\mu'(t_1) = I[E]\mu(t_1), ~~~ \mu'(t_2) = I[E]\mu(t_2),
\label{eq:9}
\vspace{-2mm}
\end{equation}
\vspace{-4mm}
\begin{equation}
f_{t_1:t_2} =  \mu'(t_2) - \mu'(t_1).
\label{eq:10}
\vspace{-1mm}
\end{equation}
Then we estimate the optical flow $F$ at pixel $p$ via by point-based $\alpha$-blending as,
\vspace{-3mm}
\begin{equation}
F = \sum_{i=1}^{N}  f_i \alpha_i \prod_{j=1}^{i-1}(1 - \alpha_j).
\label{eq:11}
\vspace{-3mm}
\end{equation}
The optical flow $\hat{\mathcal{F}}$ at time $t$ are rendered by the differentiable rendering similar as Equation~\ref{eq:8}.
In order to distill optical flow knowledge from 2D foundation model into 4D space, we adopt SEA-RAFT\cite{wang2025sea} to extract optical flow $\mathcal{F}$ as pseudo ground truth for distillation by optical flow loss,    
\vspace{-1.5mm}
\begin{equation}
\mathcal{L}_{F} = \lambda_{f} \mathcal{L}_{1}(\mathcal{F}, \hat{\mathcal{F}})
+ (1-\lambda_{f}) \mathcal{L}_{1}(\mathcal{I}_{next}, \mathcal{T}(\hat{\mathcal{I}} | \hat{\mathcal{F}} )),
\label{eq:12}
\vspace{-1.5mm}
\end{equation}
where $\lambda_{f}$ is a scale factor, and $\mathcal{I}_{next}$ denotes the image at the next time step. 
$\mathcal{T}$ is used to warp rendered image to next time step according to optical flow.   
The optical flow $\mathcal{F}$ extracted from foundation model is only required during training and is not needed during inference.

\subsection{Optimization}
All Gaussian attributes, along with the parameters of NMFF, are optimized end-to-end in a self-supervised manner. 
Meanwhile, adaptive densification and pruning strategies are introduced to enhance the fitting of the 3D scene.
The overall training loss is given as,
\vspace{-1.5mm}
\begin{equation}
\mathcal{L} = \mathcal{L}_I + \lambda_{1} \mathcal{L}_{D} + \lambda_{2} \mathcal{L}_{F} + \lambda_{3} \mathcal{L}_{sky} + \lambda_{4} \mathcal{L}_{reg},
\label{eq:16}
\vspace{-1.5mm}
\end{equation}
where $\lambda_{1,2,3,4}$ are scale factors and $\mathcal{L}_{reg}$ is the regularization term. 
The image loss $\mathcal{L}_I$ combines L1 and SSIM losses between rendered and observed images,
\vspace{-1.5mm}
\begin{equation}
\mathcal{L}_I = (1-\lambda_{ssim}) \mathcal{L}_{1}(\mathcal{I}, \hat{\mathcal{I}}) + \lambda_{ssim} \mathcal{L}_{ssim}(\mathcal{I}, \hat{\mathcal{I}}),
\label{eq:13}
\vspace{-1.5mm}
\end{equation}
where $\lambda_{ssim}$ is a scale factor.
The depth loss $\mathcal{L}_{D}$ is a L1 loss between inverse of rendered depth and generated depth by projecting sparse LiDAR points onto camera plane as,
\vspace{-1.5mm}
\begin{equation}
\mathcal{L}_{D} = \mathcal{L}_{1}(\mathcal{D}, \hat{\mathcal{D}}).
\label{eq:14}
\vspace{-1.5mm}
\end{equation}
The sky opacity loss $\mathcal{L}_{sky}$ is a binary cross entropy loss for sky refining using sky mask $M_{sky}$ from SegFormer~\cite{xie2021segformer},
\vspace{-1.2mm}
\begin{equation}
\mathcal{L}_{sky} = \mathcal{L}_{bce}(O, 1-M_{sky}),
\label{eq:15}
\vspace{-1.2mm}
\end{equation}
where the accumulated opacity denotes as $O = \sum_{i=1}^{N} \alpha_i \prod_{j=1}^{i-1}(1 - \alpha_j)$.
Note the sky mask is only required during training and is not needed during inference.
\section{Experiments}
\label{sec:experiments}
\begin{figure*}[!t]
    \centering
    \includegraphics[width=1.0\textwidth]{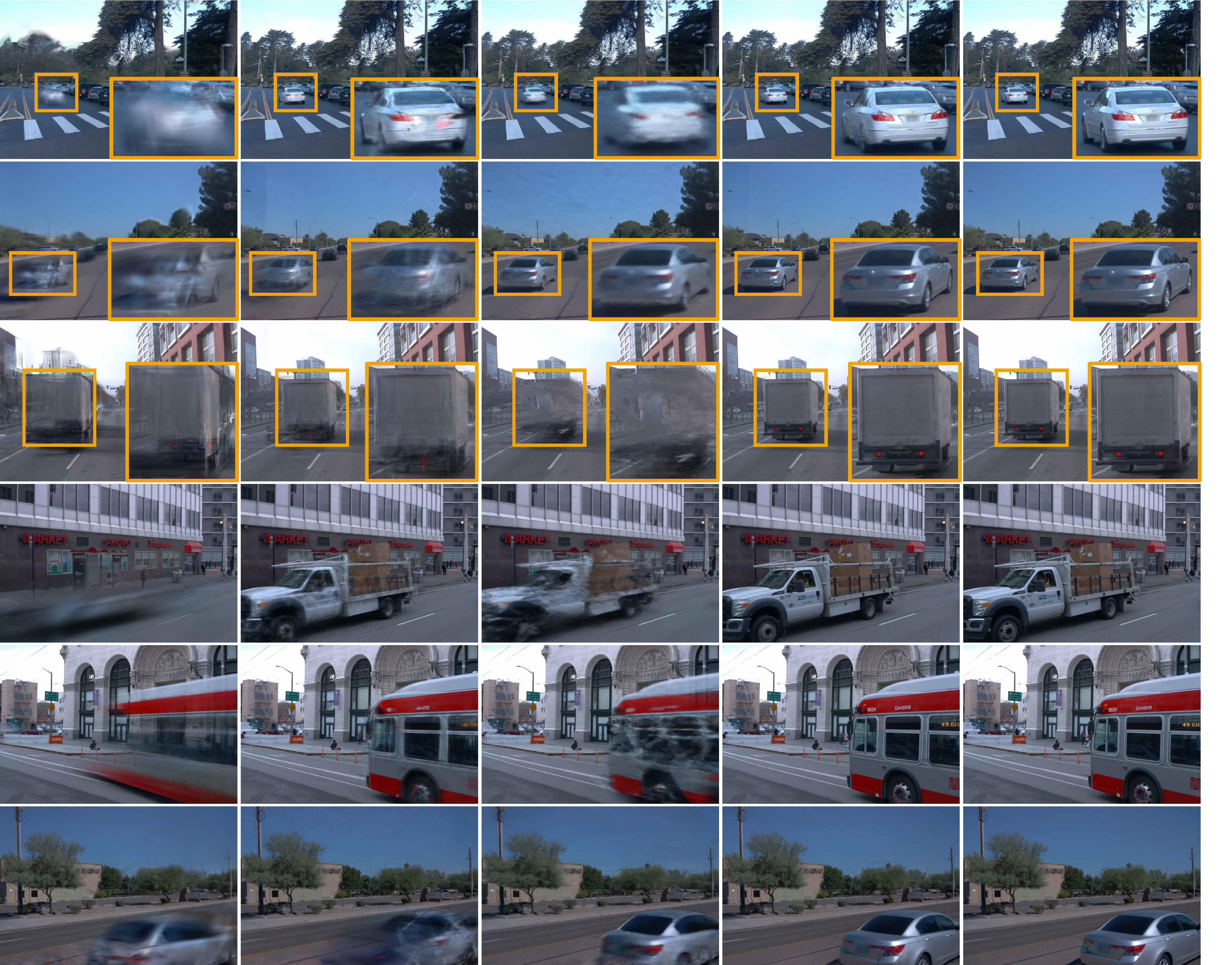}
    \vspace{-2mm} 
    \begin{minipage}{\textwidth}
    \vspace{-2mm} 
     \makebox[\textwidth][c]{%
        \hspace*{-0.3cm} 
        \parbox[t]{0.16\textwidth}{\centering \small 3D GS}
        \hspace{0.025\textwidth}
        \parbox[t]{0.16\textwidth}{\centering \small PVG}
        \hspace{0.025\textwidth}
        \parbox[t]{0.16\textwidth}{\centering \small EmerNeRF}
        \hspace{0.025\textwidth}
        \parbox[t]{0.16\textwidth}{\centering \small \textbf{SplatFlow (ours)}}
        \hspace{0.025\textwidth}
        \parbox[t]{0.16\textwidth}{\centering \small G.T.}
    }
    \end{minipage}
    \vspace{-6mm}
    \caption{Visual comparison of novel view synthesis on Waymo dataset. Bounding boxes indicate the zoomed-in dynamic areas. }
  \label{fig:waymo1}
  \vspace{-5mm}
\end{figure*}

\subsection{Datasets, Metrics and Baselines}
Following PVG~\cite{chen2023periodic}, we validate our approach on two widely-used datasets in autonomous driving: Waymo Open Dataset~\cite{sun2020scalability} and KITTI Dataset~\cite{geiger2012}. 
Both offer multi-sensor data clips including synchronized and calibrated camera and LiDAR data from urban driving scenarios. 
We train and test our methods, along with all baselines, using full-resolution images: 1920$\times$1280 for the Waymo dataset and 1242$\times$375 for the KITTI dataset.

Following prior research~\cite{turki2023suds, yang2023emernerf, zhou2024drivinggaussian, yan2024street, chen2023periodic}, we evaluate image reconstruction and novel view synthesis using three widely-adopted benchmark metrics: peak signal-to-noise ratio (PSNR), structural similarity index measure (SSIM), and learned perceptual image patch similarity (LPIPS).

We compare our method with extensive baselines including  
S-NeRF~\cite{ziyang2023snerf},
StreetSurf~\cite{guo2023streetsurf},
3DGS~\cite{kerbl20233d},
NSG~\cite{ost2021neural},
Mars~\cite{wu2023mars},
SUDS~\cite{turki2023suds},
EmerNerf~\cite{yang2023emernerf},
PVG~\cite{chen2023periodic},
StreetGaussian~\cite{yan2024street}
on the Waymo and KITTI benchmarks.
See supplementary material for more details and results.

\begin{figure}[!h]
    \centering
    \includegraphics[width=1.0\columnwidth]{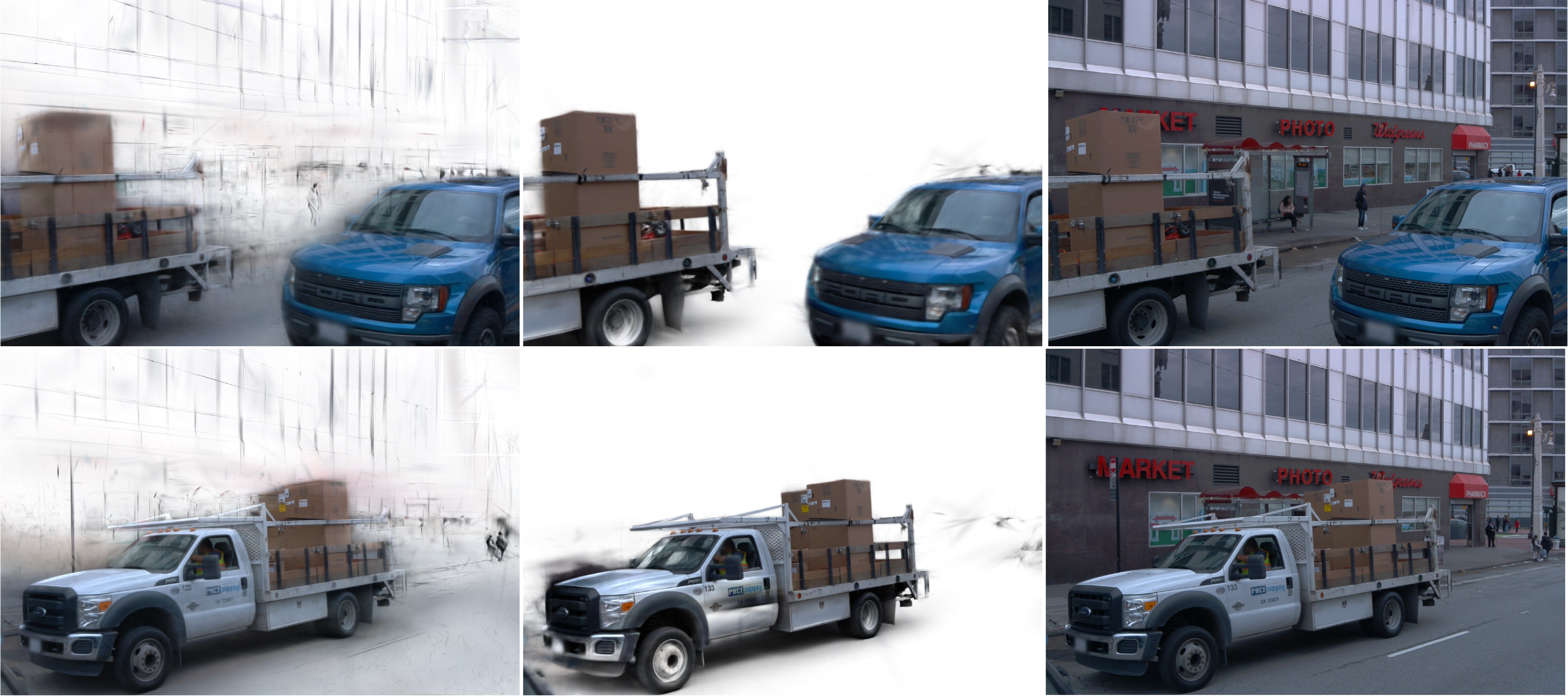}
    \vspace{-3mm} 
    \makebox[\columnwidth][c]{%
        \parbox[t]{0.13\textwidth}{\centering \footnotesize PVG}
        \hspace{0.022\textwidth}
        \parbox[t]{0.13\textwidth}{\centering \footnotesize \textbf{SplatFlow (ours)}}
        \hspace{0.022\textwidth}
        \parbox[t]{0.13\textwidth}{\centering \footnotesize Reference}
    }
    \caption{Dynamic object decomposition comparison on Waymo.}
  \label{fig:decomposition}
  \vspace{-8mm}
\end{figure}
\begin{figure*}[!t]
    \centering
    \includegraphics[width=1.0\textwidth]{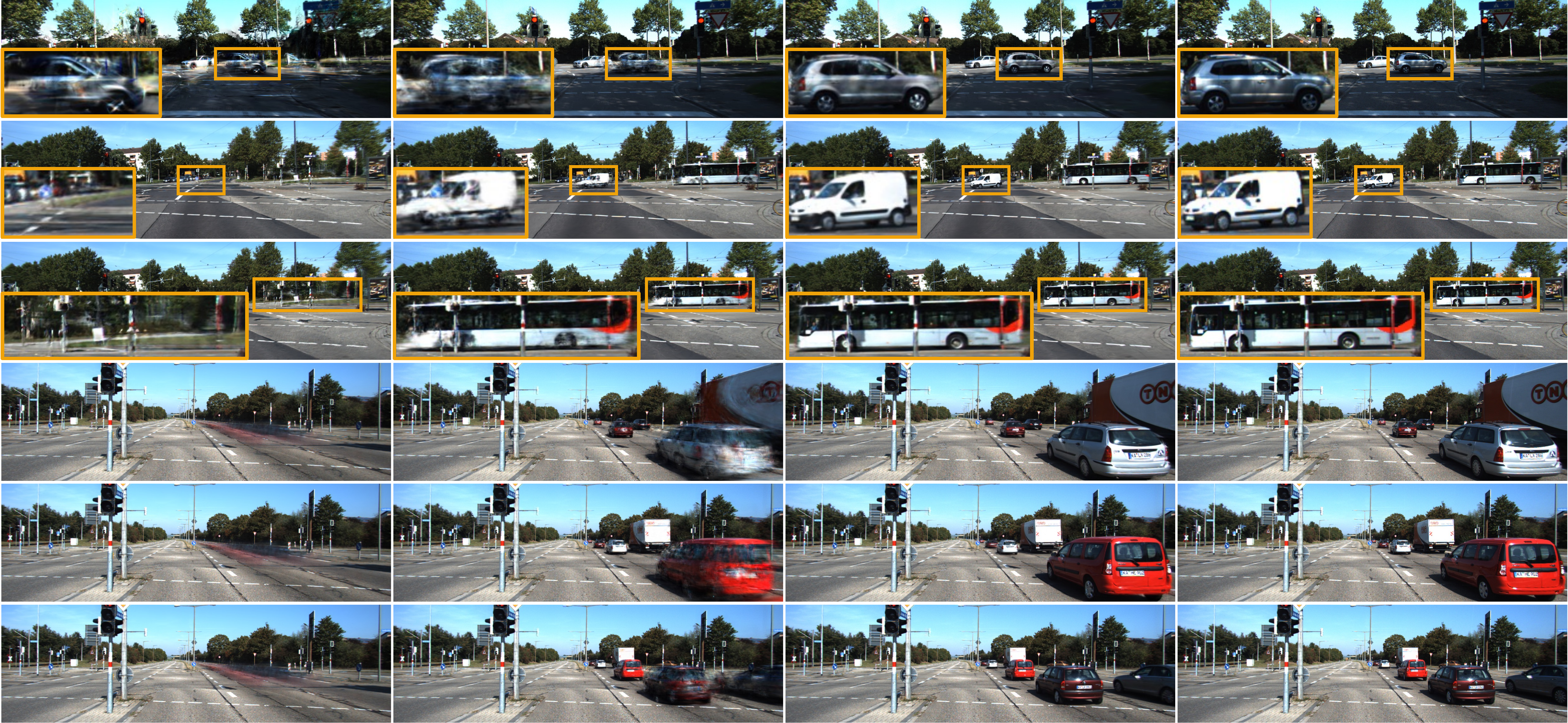}
    \vspace{-2mm}
    \begin{minipage}{\textwidth}
    \vspace{-2mm} 
    \makebox[\textwidth][c]{%
        \hspace*{-0.2cm} 
        \parbox[t]{0.17\textwidth}{\centering \small 3D GS}
        \hspace{0.08\textwidth}
        \parbox[t]{0.16\textwidth}{\centering \small PVG}
        \hspace{0.09\textwidth}
        \parbox[t]{0.17\textwidth}{\centering \small \textbf{SplatFlow (ours)}}
        \hspace{0.065\textwidth}
        \parbox[t]{0.17\textwidth}{\centering \small G.T.}
    }
    \end{minipage}
    \vspace{-6mm}
    \caption{Visual comparison of novel view synthesis on KITTI dataset. Bounding boxes indicate the zoomed-in dynamic areas.}
  \label{fig:kitti1}
  \vspace{-6mm}
\end{figure*}
\begin{table}[!h]
\centering
\resizebox{0.5\textwidth}{!}{%
\begin{tabular}{c|ccc|ccc}
\hline
\multicolumn{1}{l|}{} & \multicolumn{3}{c|}{Image Reconstruction} & \multicolumn{3}{c}{Novel View Synthesis} \\
 & PSNR$\uparrow$ & SSIM$\uparrow$ & LPIPS$\downarrow$ & PSNR$\uparrow$ & SSIM$\uparrow$ & LPIPS$\downarrow$ \\ \hline
S-NeRF~\cite{ziyang2023snerf} & 19.67 & 0.528 & 0.387 & 19.22 & 0.515 & 0.400 \\
StreetSurf~\cite{guo2023streetsurf} & 26.70 & 0.846 & 0.387 & 23.78 & 0.822 & 0.401 \\
3DGS~\cite{kerbl20233d} & 27.99 & 0.866 & 0.372 & 25.08 & 0.822 & 0.319 \\
NSG~\cite{ost2021neural} & 24.08 & 0.656 & 0.293 & 21.01 & 0.571 & 0.487 \\
Mars~\cite{wu2023mars} & 21.81 & 0.681 & 0.441 & 20.69 & 0.636 & 0.453 \\
SUDS~\cite{turki2023suds} & 28.83 & 0.805 & 0.430 & 21.83 & 0.656 & 0.405 \\
EmerNerf~\cite{yang2023emernerf} & 28.11 & 0.786 & 0.289 & 25.92 & 0.763 & 0.384 \\
PVG~\cite{chen2023periodic} & 32.46 & 0.910 & 0.373 & 28.11 & 0.849 & 0.279 \\
\hline
SplatFlow & \textbf{33.64} &\textbf{ 0.951} & \textbf{0.198} &\textbf{ 28.71} & \textbf{0.874} &\textbf{0.239} \\
\hline
\end{tabular}
}%
\vspace{-3mm}
\caption{Performance comparison on Waymo dataset.}
\label{tab:table1}
\vspace{-6mm}
\end{table}
\begin{table}[!ht]
\centering
\setlength{\tabcolsep}{4pt}
\resizebox{0.45\textwidth}{!}{
\begin{tabular}{@{}l*{6}{c}@{}}
\toprule
& \multicolumn{3}{c}{PVG~\cite{chen2023periodic}} & \multicolumn{3}{c}{SplatFlow} \\
\cmidrule(lr){2-4} \cmidrule(lr){5-7}
& PSNR$\uparrow$ & SSIM$\uparrow$ & LPIPS$\downarrow$ & PSNR$\uparrow$ & SSIM$\uparrow$ & LPIPS$\downarrow$ \\
\midrule
\addlinespace[2pt]
Seg. 1058\dots & 26.86 & 0.840 & 0.273 & \textbf{27.46} & \textbf{0.880} & \textbf{0.253} \\
Seg. 2259\dots & 26.45 & 0.824 & 0.309 & \textbf{29.30} & \textbf{0.856} & \textbf{0.290} \\
Seg. 7670\dots & 26.58 & 0.820 & 0.307 & \textbf{28.89} & \textbf{0.855} & \textbf{0.283} \\
Seg. 5083\dots & 28.35 & 0.893 & 0.213 & \textbf{30.29} & \textbf{0.928} & \textbf{0.169} \\
\addlinespace[2pt]
\midrule
& PSNR*$\uparrow$ & SSIM*$\uparrow$ & LPIPS*$\downarrow$ & PSNR*$\uparrow$ & SSIM*$\uparrow$ & LPIPS*$\downarrow$ \\
\midrule
\addlinespace[2pt]
Seg. 1058\dots &  27.42 & 0.835  &  0.267 & \textbf{28.50} & \textbf{0.877} & \textbf{0.234} \\
Seg. 2259\dots &  22.55&  0.803 & 0.350  & \textbf{31.61} & \textbf{0.899} & \textbf{0.191} \\
Seg. 7670\dots &  23.56 & 0.756 & 0.330 & \textbf{28.76} & \textbf{0.820} & \textbf{0.289} \\
Seg. 5083\dots &  25.58 &  0.814 &  0.290 & \textbf{26.71} & \textbf{0.855} & \textbf{0.211} \\
\bottomrule
\end{tabular}
}
\vspace{-3mm}
\caption{Novel view synthesis results on Waymo dataset (*denotes dynamic elements only).}
\label{tab:table2}
\vspace{-3mm}
\end{table}
\begin{table}[!h]
\centering
\resizebox{0.5\textwidth}{!}{
\begin{tabular}{c|ccc|ccc}
\hline
\multicolumn{1}{l|}{} & \multicolumn{3}{c|}{Image Reconstruction} & \multicolumn{3}{c}{Novel View Synthesis} \\
 & PSNR$\uparrow$ & SSIM$\uparrow$ & LPIPS$\downarrow$ & PSNR$\uparrow$ & SSIM$\uparrow$ & LPIPS$\downarrow$ \\ \hline
S-NeRF~\cite{ziyang2023snerf} & 19.23 & 0.664 & 0.193 & 18.71 & 0.606 & 0.352 \\
StreetSurf~\cite{guo2023streetsurf} & 24.14 & 0.819 & 0.257 & 22.48 & 0.763 & 0.304 \\
3DGS~\cite{kerbl20233d} & 21.02 & 0.811 & 0.202 & 19.54 & 0.776 & 0.224 \\
NSG~\cite{ost2021neural} & 26.66 & 0.806 & 0.186 & 21.53 & 0.673 & 0.254 \\
Mars~\cite{wu2023mars} & 27.96 & 0.900 & 0.185 & 24.23 & 0.845 & 0.160 \\
SUDS~\cite{turki2023suds} & 28.31 & 0.876 & 0.185 & 22.77 & 0.797 & 0.171 \\
EmerNeRF~\cite{yang2023emernerf} & 26.95 & 0.828 & 0.218 & 25.24 & 0.801 & 0.237 \\
PVG~\cite{chen2023periodic} & 32.83 & 0.937 & 0.070 & 27.43 & 0.896 & 0.114 \\
\hline
SplatFlow & \textbf{33.37} & \textbf{0.943} & \textbf{0.057} & \textbf{28.32} & \textbf{0.932} & \textbf{0.089} \\
\hline
\end{tabular}}
\vspace{-3mm}
\caption{Performance comparison on KITTI dataset.}
\label{tab:table3}
\vspace{-7mm}
\end{table}

\subsection{Evaluation on Waymo Open Dataset}
\begin{figure}[!h]
    \centering
    \includegraphics[width=1.0\columnwidth]{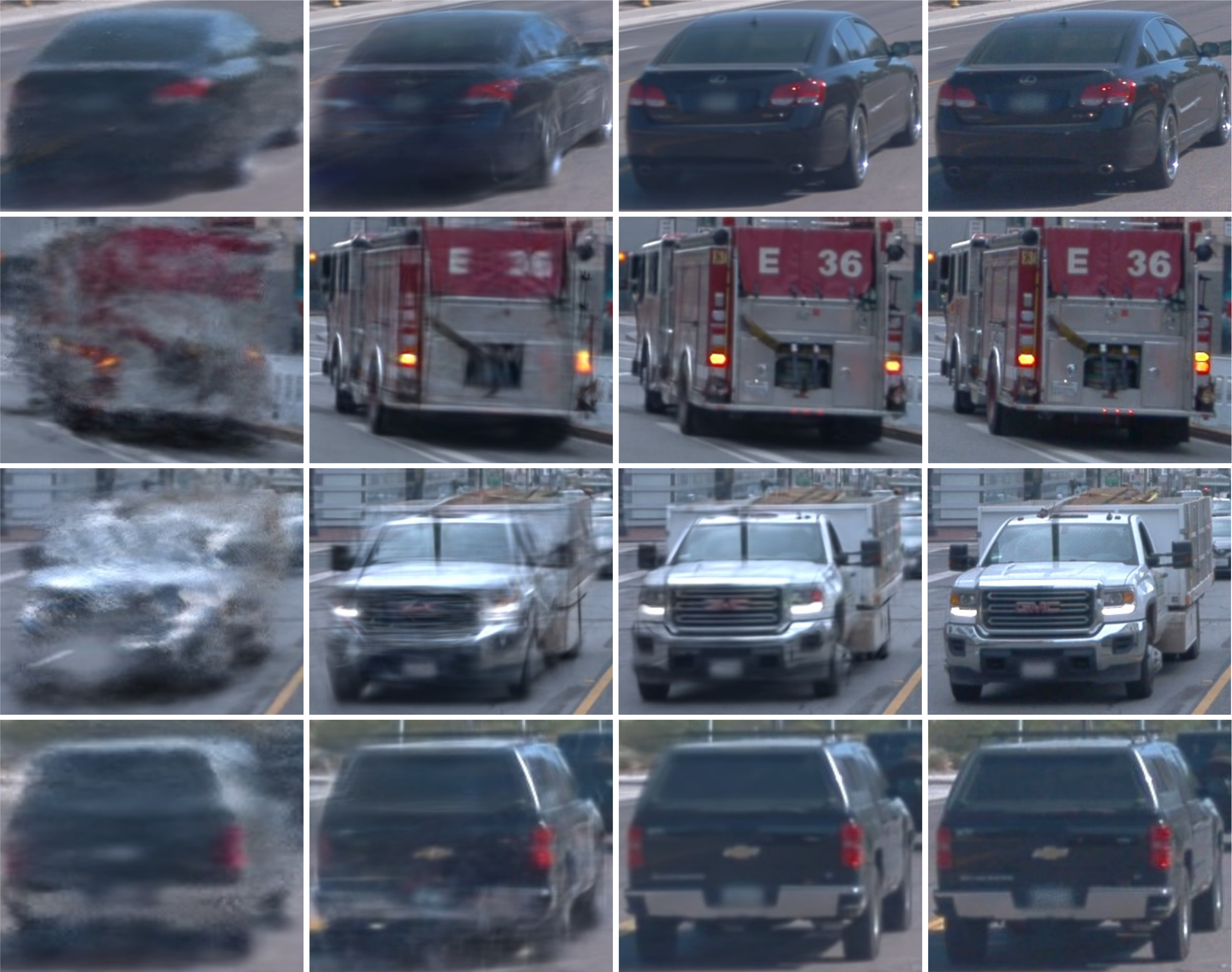}
    \vspace{-4mm}
    \begin{minipage}{\columnwidth}
        \vspace{-3mm} 
        \makebox[\columnwidth][c]{%
            \hspace*{0.1cm} 
            \parbox[t]{0.16\columnwidth}{\centering \footnotesize EmerNeRF}
            \hspace{0.06\columnwidth}
            \parbox[t]{0.17\columnwidth}{\centering \footnotesize PVG}
            \hspace{0.06\columnwidth}
            \parbox[t]{0.17\columnwidth}{\centering \footnotesize \textbf{SplatFlow}}
            \hspace{0.06\columnwidth}
            \parbox[t]{0.17\columnwidth}{\centering \footnotesize G.T.}
        }
    \end{minipage}
    \caption{Detail comparison of novel view synthesis on Waymo.}
  \label{fig:dynamic_render}
  \vspace{-4mm}
\end{figure}

Following PVG~\cite{chen2023periodic}, we evaluated our SplatFlow against baselines using the Waymo Open dataset. 
Table~\ref{tab:table1} summarizes the average metrics for both image reconstruction and novel view synthesis tasks on selected dynamic scenes. 
SplatFlow consistently outperforms all baselines across all evaluated metrics for both tasks. 
In image reconstruction, SplatFlow achieves high image quality with the highest scores: PSNR at 33.64, SSIM at 0.951, and LPIPS at 0.198. 
For novel view synthesis, SplatFlow produces high-quality renderings of unseen timestamps, reaching PSNR of 28.71, SSIM of 0.874 and LPIPS of 0.239. 
These improvements are visually validated in Fig.~\ref{fig:waymo1} and Fig.~\ref{fig:dynamic_render} (dynamic only), showing exceptional clarity in details for both static and dynamic regions.
The dynamic objects in Fig.~\ref{fig:dynamic_render} are selected from a distant background car in an extremely zoomed-in view.
Compared with baselines which show artifacts such as ghosting and blur in regions of dynamic elements, our method preserves high-quality textures and fine details, producing more accurate renderings at novel viewpoints.

\begin{table*}[!t]
    \centering
    \begin{minipage}{0.6\textwidth}
    \resizebox{1.0\textwidth}{!}{
    \begin{tabular}{l ccc ccc ccc}
    \toprule
    & \multicolumn{3}{c}{KITTI - 75\%} & \multicolumn{3}{c}{KITTI - 50\%} & \multicolumn{3}{c}{KITTI - 25\%} \\
    \cmidrule(lr){2-4} \cmidrule(lr){5-7} \cmidrule(lr){8-10}
    & PSNR$\uparrow$ & SSIM$\uparrow$ & LPIPS$\downarrow$ & PSNR$\uparrow$ & SSIM$\uparrow$ & LPIPS$\downarrow$ & PSNR$\uparrow$ & SSIM$\uparrow$ & LPIPS$\downarrow$ \\
    \midrule
    NeRF~\cite{mildenhall2021nerf} & 18.56 & 0.557 & 0.554 & 19.12 & 0.587 & 0.497 & 18.61 & 0.570 & 0.510 \\
    NSG~\cite{ost2021neural} & 21.53 & 0.673 & 0.254 & 21.26 & 0.659 & 0.266 & 20.00 & 0.632 & 0.281 \\
    SUDS~\cite{turki2023suds} & 22.77 & 0.797 & 0.171 & 23.12 & 0.821 & 0.135 & 20.76 & 0.747 & 0.198 \\
    MARS~\cite{wu2023mars} & 24.23 & 0.845 & 0.160 & 24.00 & 0.801 & 0.164 & 23.23 & 0.756 & 0.177\\
    3DGS~\cite{kerbl20233d} & 19.19 & 0.737 & 0.172 & 19.23 & 0.739 & 0.174 & 19.06 & 0.730 & 0.180 \\
    PVG~\cite{chen2023periodic} & 27.43 & 0.896 & 0.114 & 25.92 & 0.882 & 0.114 & 22.55 & 0.833 & 0.151  \\
    StreetGS~\cite{yan2024street} & 25.79 & 0.844 & \textbf{0.081} & 25.52 & 0.841 & \textbf{0.084} & 24.53 & 0.824 & \textbf{0.090} \\
    \hline
    SplatFlow & \textbf{28.32} & \textbf{0.932} & 0.089 & \textbf{27.90} & \textbf{0.927} & 0.093 & \textbf{26.10} & \textbf{0.890} & 0.120 \\
    \bottomrule
    \end{tabular}}
    \vspace{-3mm}
    \caption{Performance comparison of novel view synthesis on KITTI-75\%, 50\% and 25\% dataset.}
    \label{tab:table4}
    \end{minipage}
    \begin{minipage}{0.35\textwidth}
    \centering
    \resizebox{1.0\textwidth}{!}{
    \begin{tabular}{cccc}
    \hline
    & PSNR$\uparrow$ & SSIM$\uparrow$ & LPIPS$\downarrow$  \\ \hline
    w/o NMFF prior & 27.69 & 0.863 & 0.282 \\
    w/o NMFF optimization & 28.14 &  0.874  & 0.269  \\
    w/o optical flow distillation & 28.28 & 0.877 & 0.252 \\
    Full & \textbf{28.99} &\textbf{0.880} & \textbf{0.249} \\ \hline
    & PSNR*$\uparrow$ & SSIM*$\uparrow$ & LPIPS*$\downarrow$  \\ \hline
    w/o NMFF prior & 27.28 & 0.827 & 0.317 \\
    w/o NMFF optimization & 27.97 & 0.843  & 0.254\\
    w/o optical flow distillation  &28.51  & 0.861 & 0.232\\
    Full & \textbf{28.90} & \textbf{0.863} & \textbf{0.231} \\ \hline
    \end{tabular}}
    \vspace{-3mm}
    \caption{Ablation study on Waymo dataset (*denotes dynamic elements only).}
    \label{tab:table5}
    \end{minipage}
    \vspace{-6mm}
\end{table*}
\begin{figure}[h]
    \centering
    \includegraphics[width=1.0\columnwidth]{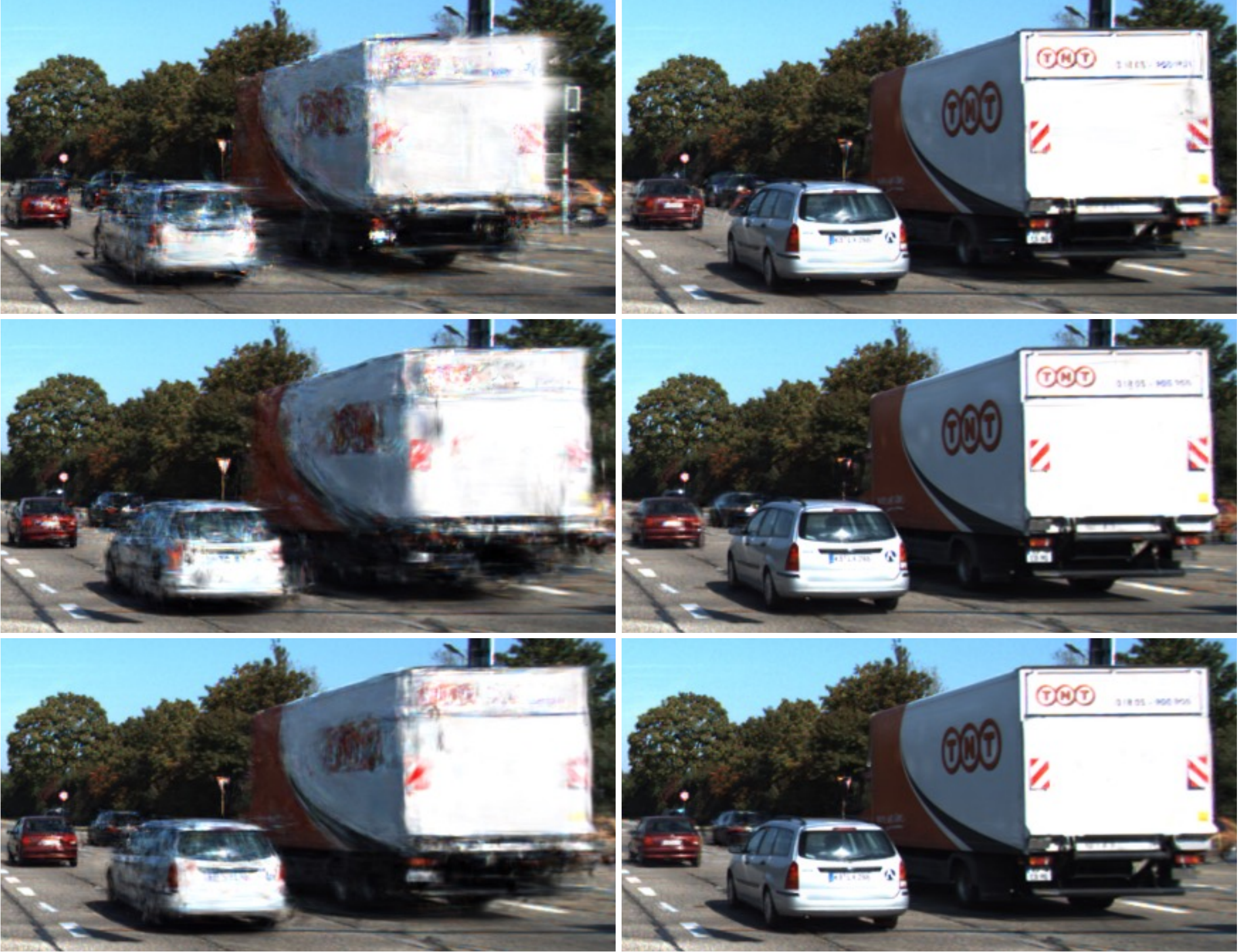}
    \vspace{-4mm}
    \begin{minipage}{\columnwidth}
    \vspace{-2mm}
    \makebox[\columnwidth][c]{%
        \parbox[t]{0.45\textwidth}{\centering \footnotesize PVG}
        \hspace{0.055\textwidth}
        \parbox[t]{0.45\textwidth}{\centering \footnotesize \textbf{SplatFlow (ours)}}
    }
    \vspace{-3mm}
    \end{minipage}
    \caption{Visual comparison of novel view synthesis on KITTI\-25\% (row1), 50\% (row2), and 75\% (row3) dataset. 
    }
  \label{fig:kitti_ablation}
  \vspace{-8mm}
\end{figure}

To further evaluate the rendering quality of dynamic objects against the primary baseline PVG, we selected additional four scenes featuring a higher presence of dynamic objects for further comparison. 
Table~\ref{tab:table2} displays the metrics for novel view synthesis, evaluated both on entire scenes and exclusively on dynamic objects (denoted as $*$). 
SplatFlow consistently surpasses PVG in all selected scenes, demonstrating higher PSNR and SSIM values and lower LPIPS scores, underscoring its effectiveness in handling complex, dynamic content.
The dynamic objects masks are obtained from 2D ground truth bounding boxes following~\cite{yang2023emernerf}. 
The PSNR* of dynamic regions on novel view synthesis significantly outperforms PVG both quantitatively and qualitatively, especially in scenes with fast ego-motion and moving objects in Seg.~2259 and Seg.~7670.
We also present the visual comparison of dynamic object decomposition with PVG in Fig.~\ref{fig:decomposition}, which illustrates that SplatFlow renders sharper and clearer image of dynamic objects.
\begin{figure}[h!]
    \centering
    \begin{minipage}{\columnwidth}
        \centering
        \includegraphics[width=1\columnwidth]{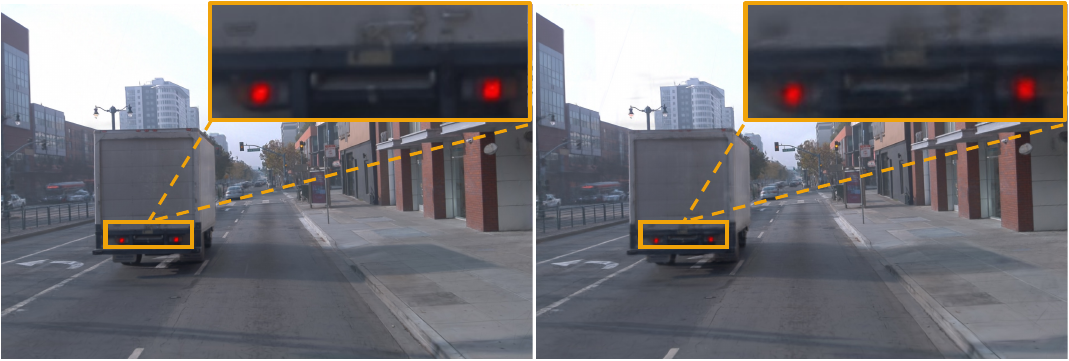}
        \raisebox{2mm}{\parbox[t]{0.3\textwidth}{\centering \footnotesize \textbf{Ours (Full)}}}
        \hspace{0.2\textwidth}
        \raisebox{2mm}{\parbox[t]{0.3\textwidth}{\centering \footnotesize  w/o Opt. flow}}
    \end{minipage}
    \begin{minipage}{\columnwidth}
        \centering
        \vspace{-1.7mm}
        \includegraphics[width=1\columnwidth]{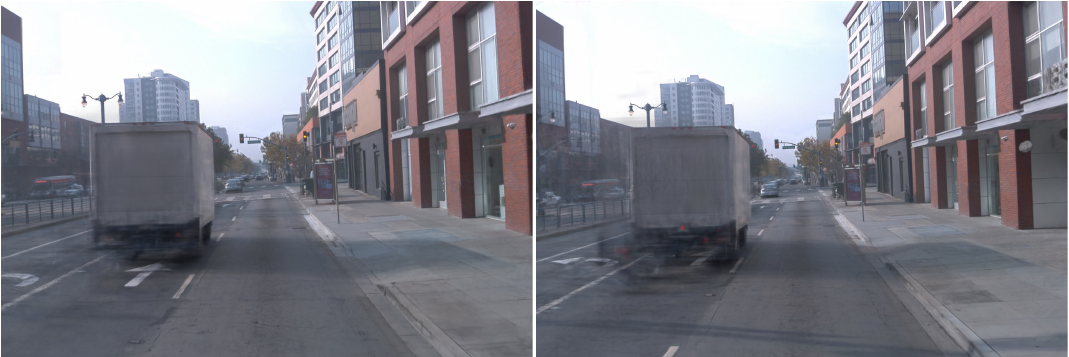}
        \raisebox{1.5mm}{\parbox[t]{0.3\textwidth}{\centering \footnotesize w/o NMFF optim.}}
        \hspace{0.2\textwidth}
        \raisebox{1.5mm}{\parbox[t]{0.3\textwidth}{\centering \footnotesize w/o NMFF prior}}
    \end{minipage}
    \vspace{-4mm}
    \caption{Visual comparison of ablation study on Waymo dataset.}
    \vspace{-4mm}
\end{figure}

\vspace{-2mm}
\subsection{Evaluation on KITTI Dataset}
\vspace{-1mm}
Following PVG~\cite{chen2023periodic}, we further evaluated our SplatFlow method in comparison to the baselines on KITTI dataset.
Table~\ref{tab:table3} presents the average metrics for both image reconstruction and novel view synthesis tasks on selected dynamic scenes characterized by extensive movement. 
SplatFlow outperforms all other methods across all metrics in both tasks, with particularly high scores in PSNR. SSIM and LPIPS, achieving 33.37, 0.943 and 0.057 for image reconstruction and 28.32, 0.932 and 0.089 for novel view synthesis. 
SplatFlow achieves exceptional results in novel view synthesis, as shown in Fig.~\ref{fig:kitti1}. Our approach generates high-fidelity renderings that preserve fine details while accurately reconstructing dynamic areas where existing approaches struggle to maintain stability.

To evaluate robustness to training data size, we follow the settings of \cite{yan2024street, wu2023mars} to evaluate our method using different train/test split configurations on the KITTI dataset. 
Table~\ref{tab:table4} and Fig.~\ref{fig:kitti_ablation} provide the performance comparisons and visualizations. 
SplatFlow consistently outperforms all other methods across all dataset subsets, achieving the highest scores. 
Even with reduced training data, SplatFlow robustly delivers high-quality novel view synthesis across varying levels of data availability, highlighting its efficient use of available information and strong generalization capabilities.
\subsection{Ablation Study}
To demonstrate the effectiveness of each component in SplatFlow, we performed ablation studies on the Waymo dataset. 
Our main contributions, NMFF prior, NMFF optimization, and optical flow distillation are analyzed in this study to assess their individual impact.
We trained four different variations: 
1) SplatFlow without pretraining on LiDAR data;
2) SplatFlow without optimization with 4D Gaussians on image data;
3) SplatFlow without optical flow distillation from foundational model;
4) SplatFlow full method.
Table~\ref{tab:table5} presents the average evaluation metrics across test scenes used in Table~\ref{tab:table2}, showing results for both entire scenes and dynamic objects (indicated as $*$).
Ablation study validates the contribution of each component to the overall performance.

\section{Conclusion}
\label{sec:conclusion}
In this paper, we propose SplatFlow, a novel self-supervised Dynamic Gaussian Splatting within Neural Motion Flow Fields (NMFF) for accurate reconstruction and real-time rendering in dynamic urban scenarios.
The core idea of SplatFlow is to seamlessly integrate time-dependent 4D Gaussian representations within NMFF in a unified framework, where NMFF implicitly models the motions of dynamic components across time through self-supervision.
Experimental evaluations show that SplatFlow outperforms state-of-the-art across all standard metrics on the Waymo Open and KITTI datasets, without requiring expensive annotations of dynamic object detection and tracking.
\clearpage
\setcounter{page}{1}
\maketitlesupplementary
In this supplementary material, we provide more implementation details of SplatFlow in Appendix~\ref{sec:implementation}. 
We provide more visualization of 3D LiDAR points within NMFF in Appendix~\ref{sec:visualization_lidar}. 
We present more detailed comparison visualizations of dynamic object synthesis from novel views in Appendix~\ref{sec:visualization}.
We provide more visualizations of dynamic objects decomposition in Appendix~\ref{sec:visualization_dynamic_object}.
We present visualizations of the novel view synthesis on newly generated ego-car trajectories in Appendix~\ref{sec:visualization_new_trajectory}.
We include visualizations of rendered RGB image, optical flow and depth in Appendix~\ref{sec:visualization_flow}.
Runtime performance comparisons are provided in Appendix~\ref{sec:running time}.
Video demonstrations are included in Appendix~\ref{sec:demo}.

\section{Implementation Details}
\label{sec:implementation}
Each field in the Neural Motion Flow Field~(NMFF) consists of eight ReLU-MLP stacks. 
All MLPs are followed by a ReLU activation, except for the final prediction layer, where the middle hidden dimensions are configured as 128. 

For NMFF pre-training, we follow the approach in~\cite{pontes2020scene} to generate pseudo scene flow labels, excluding ground points from the Waymo and KITTI datasets. 
The raw 3D LiDAR points are utilized without cropping to a smaller range. 
We use a learning rate of 8e-3 with the Adam optimizer, optimizing each scene for up to 4000 iterations with early stopping.
Additionally, point cloud densification is performed by accumulating point clouds through Euler integration, using per-pair scene flow estimations.

During the 4D Gaussian with NMFF optimization, we configure the position learning rate to a range from 1.6e-5 to 1.6e-6, the opacity learning rate to 0.05, the scale learning rate to 0.005, the feature learning rate to 2.5e-3, and the rotation learning rate to 0.001. 
The intervals for densification and opacity reset are set to 500 and 3000, respectively. 
We set the densify gradient threshold for decomposed static and dynamic Gaussians as 1.7e-4 and 1e-4, respectively. 
The Spherical Harmonics degree for each Gaussian is set to 3. 
For NMFF optimization, we set the learning rate to 1e-4.
For training losses, we use coefficients $\lambda_{1} = 0.1$, $\lambda_{2} = 0.005$, $\lambda_{3} = 0.05$, $\lambda_{4} = 0.001$, $\lambda_{ssim} = 0.2$ and $\lambda_{f} = 0.8$.
\section{Visualization of LiDAR Points in NMFF}
\label{sec:visualization_lidar}
\begin{figure}[!h]
    \centering
    \includegraphics[width=1.0\columnwidth]{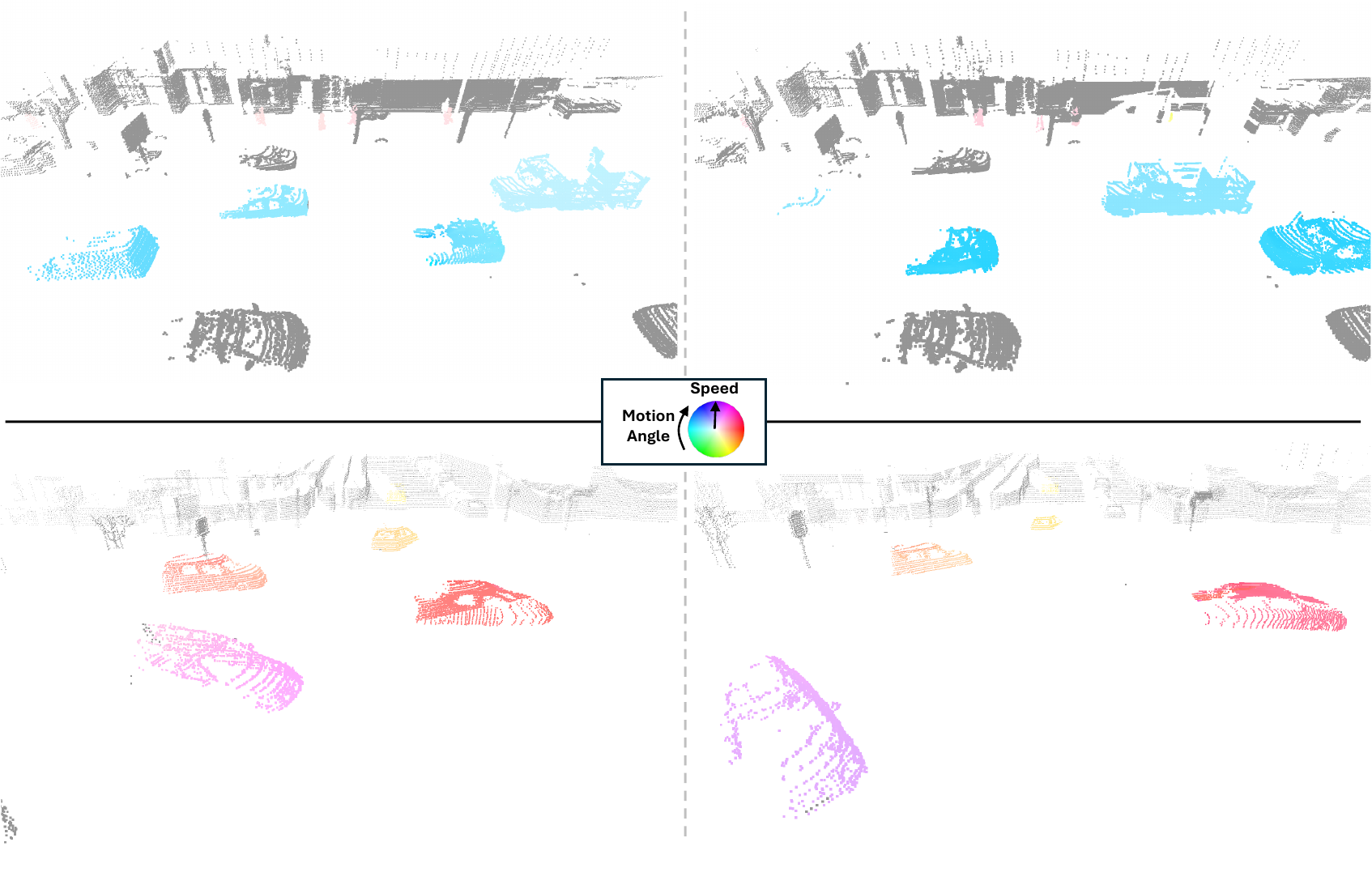}
    \vspace{-2mm} 
    \makebox[\columnwidth][c]{%
        \parbox[t]{0.15\textwidth}{\centering \footnotesize t}
        \hspace{0.055\textwidth}
        \parbox[t]{0.15\textwidth}{\centering \footnotesize t+10}
    }
    \caption{Visualization of 3D LiDAR points within NMFF on Waymo dataset.}
  \label{fig:flow_zoom}
\end{figure}

We provide more visualization of 3D LiDAR points within NMFF on the Waymo dataset in Fig.~\ref{fig:flow_zoom}.
The color wheel in the center represents the flow magnitude through color intensity and the flow direction via angle. 
As illustrated, NMFF accurately predicts the 3D motion flow of 3D LiDAR points for dynamic objects in driving scenarios.

\vspace{-3mm}
\begin{figure}[!h]
    \centering
    \includegraphics[width=1.0\columnwidth]{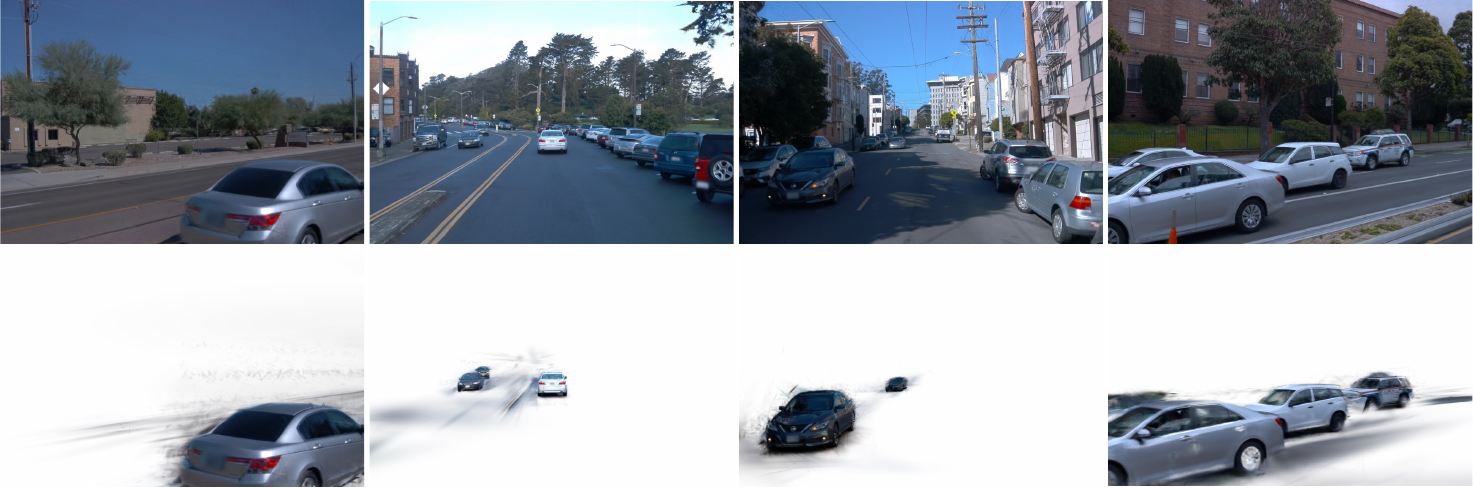}
    \vspace{-5mm}
    \caption{Dynamic object decomposition results of SplatFlow on Waymo. Row1: Rendered scene, Row2: Corresp. Decomposition}
    \label{fig:decomposition}
    \vspace{-3mm}
\end{figure}

\begin{figure*}[!h]
    \centering
    \includegraphics[width=01.0\textwidth]{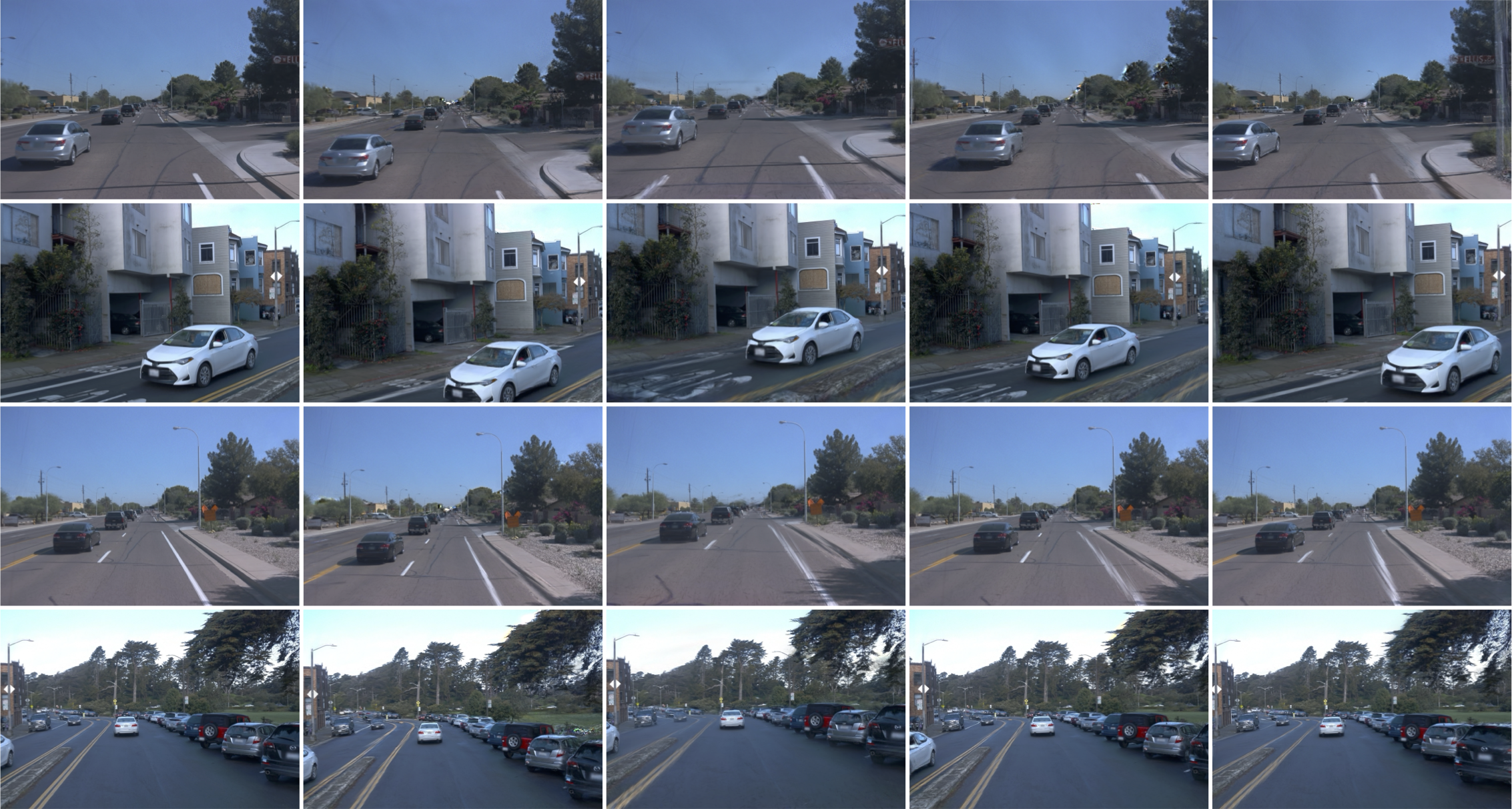}
    \vspace{-2mm} 
    \begin{minipage}{\textwidth}
    \vspace{-2mm} 
     \makebox[\textwidth][c]{%
        \parbox[t]{0.16\textwidth}{\centering \small Reference View}
        \hspace{0.025\textwidth}
        \parbox[t]{0.16\textwidth}{\centering \small Shifted Up 0.5m}
        \hspace{0.03\textwidth}
        \parbox[t]{0.16\textwidth}{\centering \small Shifted Down 0.5m}
        \hspace{0.04\textwidth}
        \parbox[t]{0.16\textwidth}{\centering \small Shifted Left 0.5m}
        \hspace{0.015\textwidth}
        \parbox[t]{0.16\textwidth}{\centering \small Shifted Right 0.5m}
    }
    \end{minipage}
    \caption{Novel view synthesis on newly generated ego-car trajectories on Waymo dataset.}
    \label{fig:waymo_nvs}
    \vspace{-4mm}
\end{figure*}
\begin{table}[!h]
\centering
\resizebox{0.3\textwidth}{!}{
\begin{tabular}{c|c|c}
\hline
& Waymo & KITTI \\
& FPS & FPS \\ \hline
S-NeRF~\cite{ziyang2023snerf} & 0.0014 & 0.0075 \\
StreetSurf~\cite{guo2023streetsurf} & 0.097 & 0.37 \\
NSG~\cite{ost2021neural} & 0.032 & 0.19 \\
Mars~\cite{wu2023mars} & 0.030 & 0.31  \\
SUDS~\cite{turki2023suds} & 0.008 & 0.04 \\
EmerNerf~\cite{yang2023emernerf} & 0.053 & 0.28\\
3DGS~\cite{kerbl20233d} & \textbf{63} & \textbf{125} \\
PVG~\cite{chen2023periodic} & 50 & 59 \\
\hline
SplatFlow & 40 & 44 \\
\hline
\end{tabular}
}
\vspace{-3mm}
\caption{The comparison running-time analysis on Waymo and KITTI datasets.}
\label{tab:run tiem}
\vspace{-3mm}
\end{table}
\section{Visual Comparison of Dynamic Object Synthesis}
\label{sec:visualization}
\begin{figure*}[!th]
    \centering
    \includegraphics[width=1.0\textwidth]{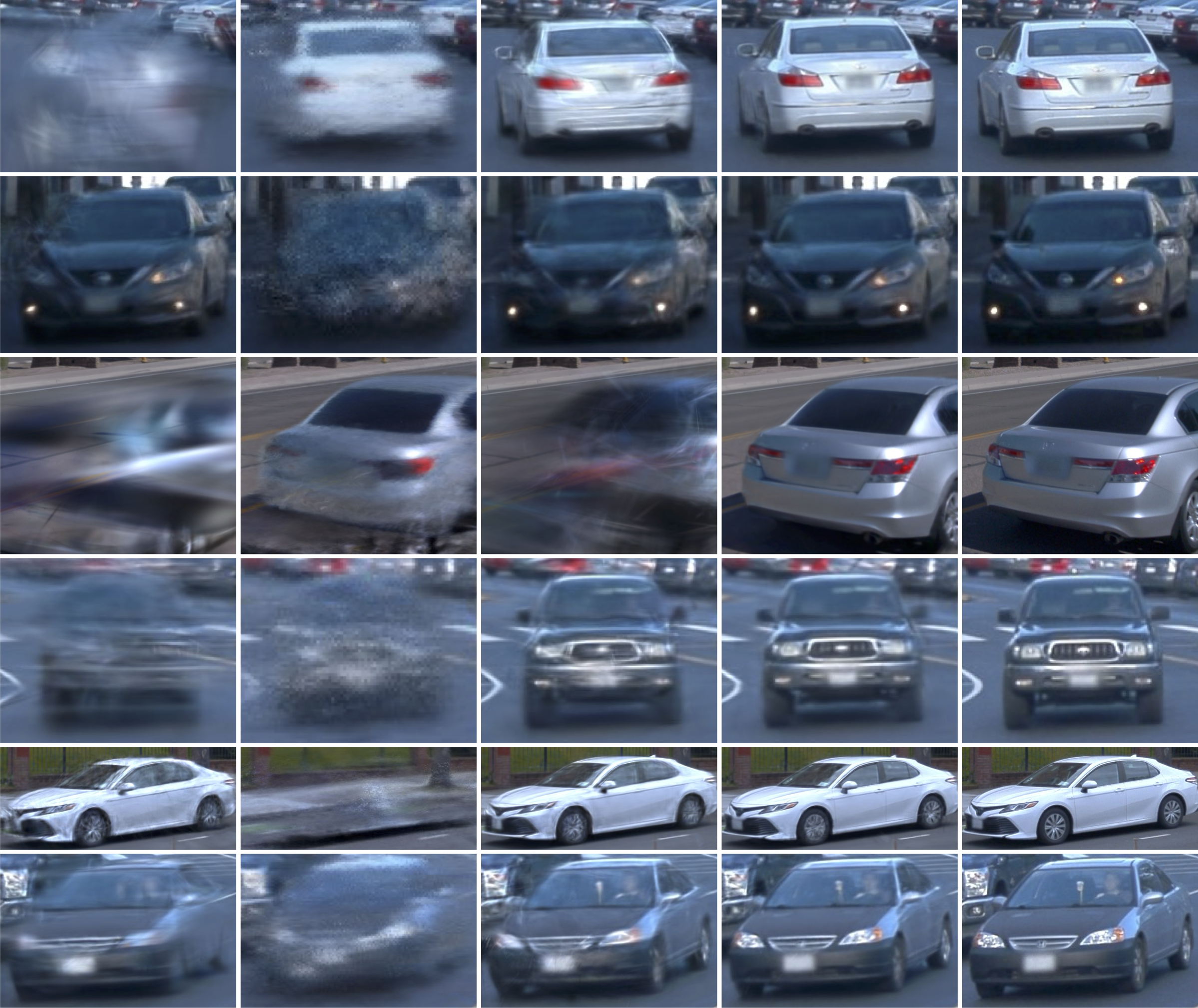}
    \vspace{-2mm} 
    \begin{minipage}{\textwidth}
    \vspace{-2mm} 
     \makebox[\textwidth][c]{%
        \parbox[t]{0.16\textwidth}{\centering \small 3D GS}
        \hspace{0.025\textwidth}
        \parbox[t]{0.16\textwidth}{\centering \small EmerNeRF }
        \hspace{0.03\textwidth}
        \parbox[t]{0.16\textwidth}{\centering \small PVG}
        \hspace{0.04\textwidth}
        \parbox[t]{0.16\textwidth}{\centering \small \textbf{SplatFlow (ours)}}
        \hspace{0.015\textwidth}
        \parbox[t]{0.16\textwidth}{\centering \small G.T.}
    }
    \end{minipage}
    \vspace{-5mm} 
    \caption{ Detailed comparison of dynamic object synthesis from novel views on Waymo dataset.}
  \label{fig:detail_waymo}
\end{figure*}

\begin{figure*}[!h]
    \centering
    \includegraphics[width=0.97\textwidth]{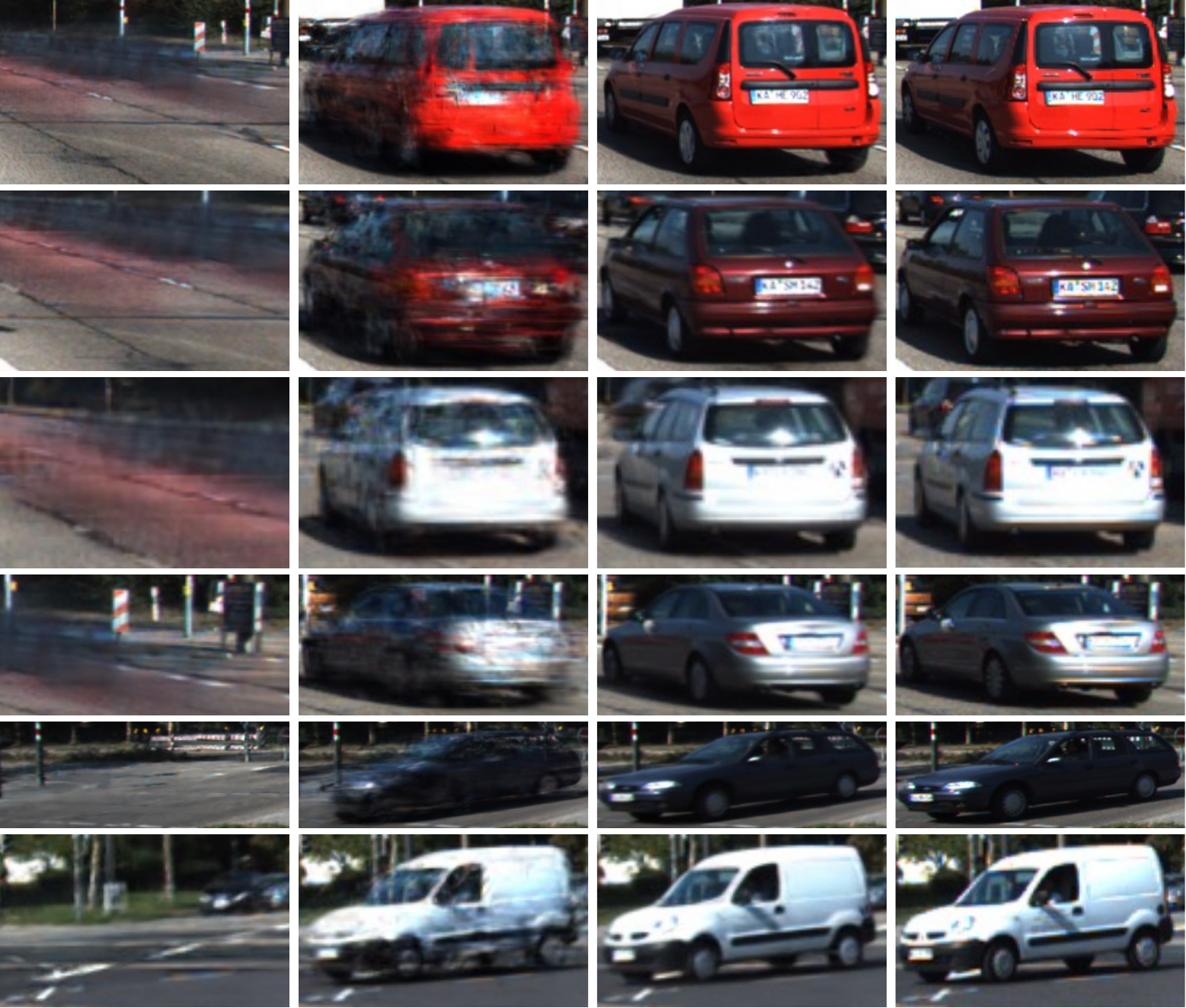}
    \vspace{-2mm} 
    \begin{minipage}{\textwidth}
    \vspace{-2mm} 
     \makebox[\textwidth][c]{%
        \parbox[t]{0.16\textwidth}{\centering \small 3D GS}
        \hspace{0.07\textwidth}
        \parbox[t]{0.16\textwidth}{\centering \small PVG}
        \hspace{0.08\textwidth}
        \parbox[t]{0.16\textwidth}{\centering \small \textbf{SplatFlow (ours)}}
        \hspace{0.07\textwidth}
        \parbox[t]{0.16\textwidth}{\centering \small G.T.}
    }
    \end{minipage}
    \vspace{-5mm} 
    \caption{ Detailed comparison of dynamic object synthesis from novel views on KITTI dataset.}
  \label{fig:detail_kitti}
\end{figure*}

We present more visual comparison details of dynamic object synthesis from novel views, showcasing results on the Waymo dataset in Fig.~\ref{fig:detail_waymo} and on the KITTI dataset in Fig.~\ref{fig:detail_kitti}.
The dynamic objects in these figures are selected from a distant background car in an extremely zoomed-in view.
As shown, SplatFlow generates sharper images with fewer blurred artifacts, particularly for high-speed vehicles, compared to the baselines.

\section{Visualization of Dynamic Object Decomposition}
\label{sec:visualization_dynamic_object}
We provide more dynamic object decomposition visualization of SlatFlow in Fig.\ref{fig:decomposition}.
As shown, SplatFlow clearly separates the dynamic objects from the rendered scenes.
\section{Visualization of Novel View Synthesis on newly Ego-car Trajectory}
\label{sec:visualization_new_trajectory}
We provide novel view synthesis in a more challenging rendering scenario on the Waymo dataset, using newly generated ego-car trajectories. 
As shown in Fig.~\ref{fig:waymo_nvs}, the results are produced from an ego-car trajectory that is shifted 0.5 meters higher, 0.5 meters lower, 0.5 meters to the left, and 0.5 meters to the right of the original trajectory. 
As demonstrated, SplatFlow can render high-quality novel images for these newly generated ego-car trajectories.

\section{Rendered Depth and Flow Visualization} 
\label{sec:visualization_flow}
\begin{figure*}[!h]
    \centering
    \includegraphics[width=1.0\textwidth]{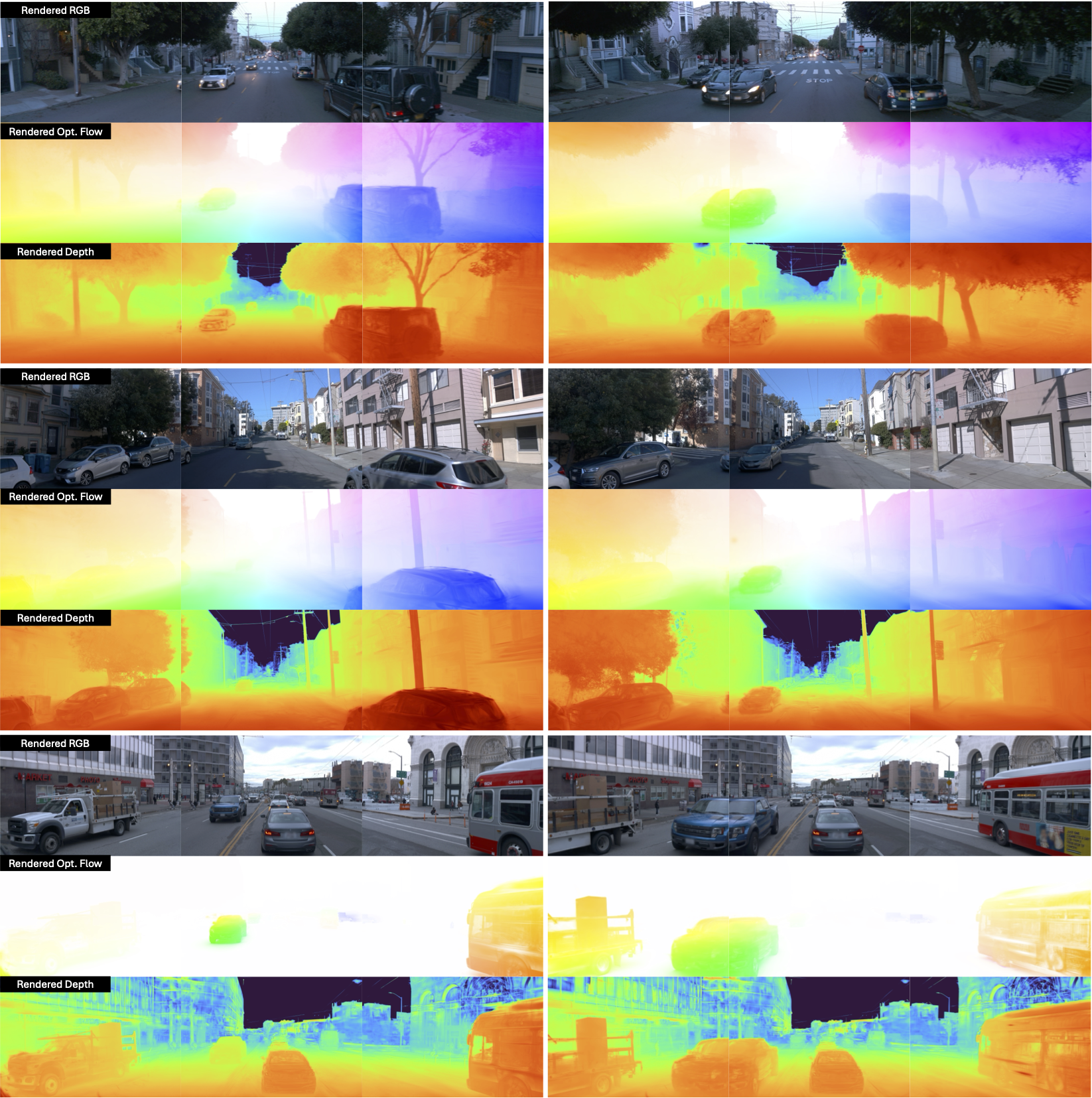}
    \vspace{-6mm}
    \caption{Visualization of rendered RGB image, optical flow, and depth by SlpatFlow on Waymo dataset.}
  \label{fig:render_waymo1}
  \vspace{-6mm}
\end{figure*}

\begin{figure*}[!h]
    \centering
    \includegraphics[width=1.0\textwidth]{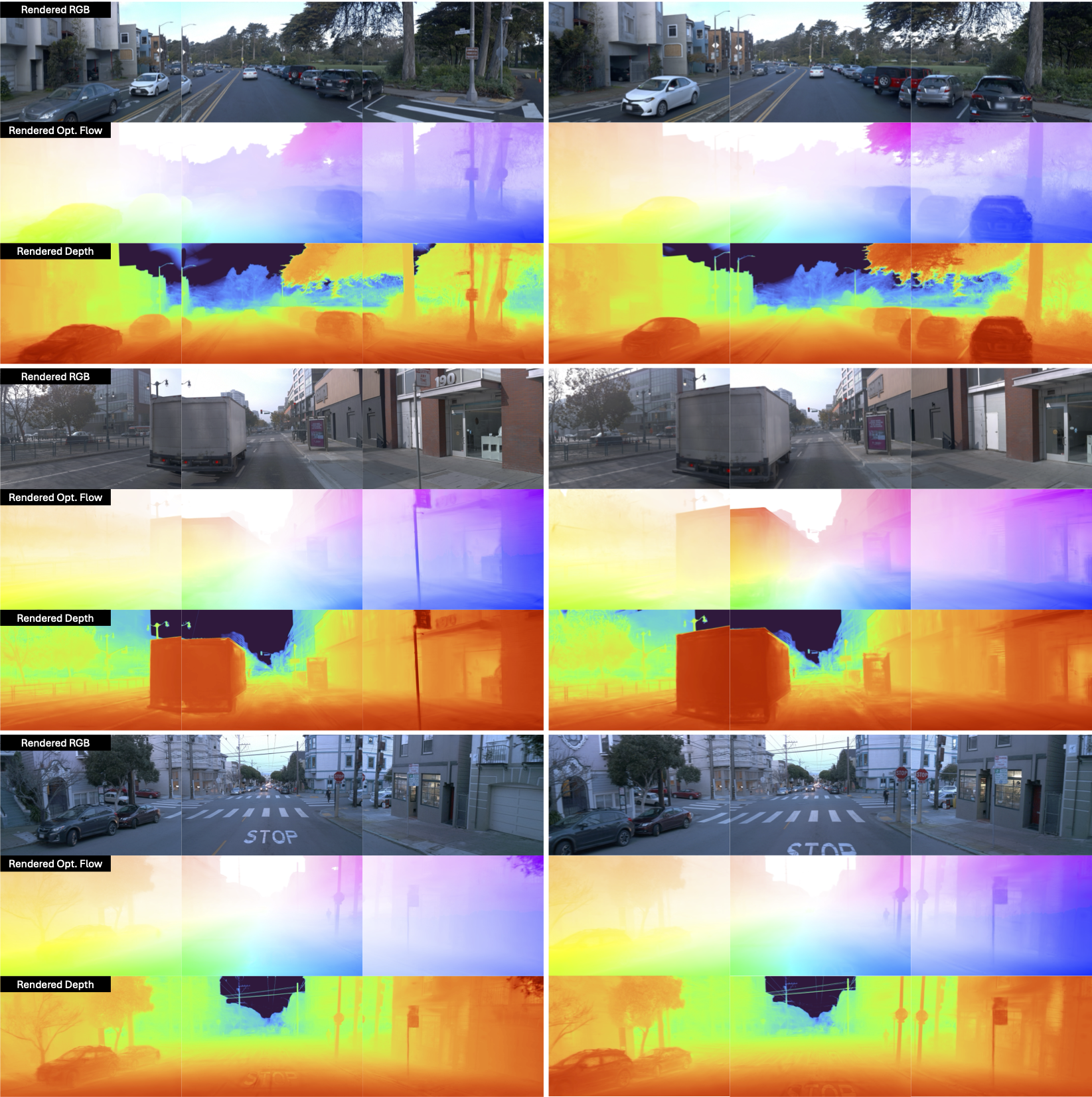}
    \vspace{-6mm}
    \caption{Visualization of rendered RGB image, optical flow, and depth by SplatFlow on Waymo dataset.}
  \label{fig:render_waymo2}
  \vspace{-6mm}
\end{figure*}

\begin{figure*}[!h]
    \centering
    \includegraphics[width=1.0\textwidth]{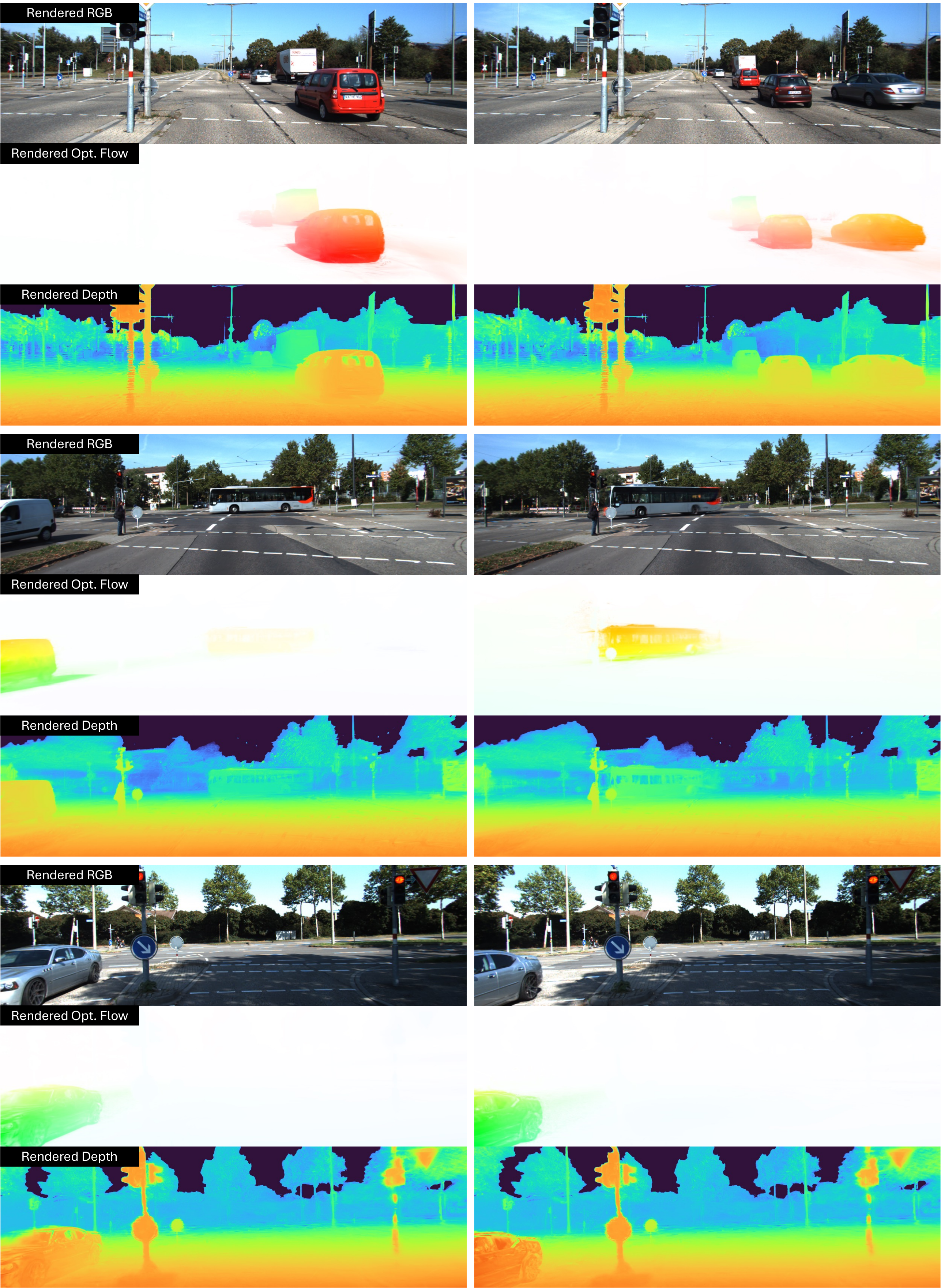}
    \vspace{-6mm}
    \caption{Visualization of rendered RGB image, optical flow, and depth by SlpatFlow on KITTI dataset.}
  \label{fig:render_kitti}
  \vspace{-6mm}
\end{figure*}

We also present visualization of rendered RGB image, optical flow, and depth on the Waymo dataset in Fig.~\ref{fig:render_waymo1} and \ref{fig:render_waymo2},  and on the KITTI dataset in Fig.~\ref{fig:render_kitti}.
In these Figures, the first row displays the rendered RGB images, the second row shows the rendered optical flow, and the third row presents the rendered depth, all generated by SplatFlow.
As shown, our SplatFlow renders sharp, clear, and dense optical flow and depth images in dynamic driving scenarios.

\section{Running-time Analysis}
\label{sec:running time}
We compare the runtime performance of SplatFlow with various baseline methods on the Waymo and KITTI datasets, as summarized in Table~\ref{tab:run tiem}.
Utilizing a single NVIDIA GeForce A6000, SplatFlow achieves real-time rendering speeds for high-resolution images after quantization and pruning optimization, delivering approximately 40 FPS at 1920$\times$1280 resolution on the Waymo dataset and around 44 FPS at 1242$\times$375 resolution on the KITTI dataset. 
Compared to NeRF-based methods such as S-NeRF~\cite{ziyang2023snerf}, StreetSurf~\cite{guo2023streetsurf}, NSG~\cite{ost2021neural}, SUDS~\cite{turki2023suds}, EmerNerf~\cite{yang2023emernerf}, SplatFlow significantly surpasses the speed of these methods, delivering real-time performance.
Compared to GS-based methods such as 3DGS~\cite{kerbl20233d} and PVG~\cite{chen2023periodic}, SplatFlow achieves higher accuracy in dynamic object rendering while maintaining efficient performance.

\section{Video Demos}
\label{sec:demo}
We include five video demos in the supplementary material. See our project page: \url{https://sites.google.com/view/splatflow}

Demos 1 and 2 showcase the results of image synthesis from novel views produced by SplatFlow, alongside baseline methods and ground truth (G.T.) data, in dynamic driving scenarios from the Waymo and KITTI datasets respectively. 
In these videos, the first and second rows display the surrounding images rendered by the baselines: 3D-GS~\cite{kerbl20233d}, PVG~\cite{chen2023periodic}, or EmerNeRF~\cite{yang2023emernerf}. 
The third row presents the surrounding images rendered by our SplatFlow, while the final row shows the G.T. surrounding images. 
To provide a clearer comparison in the visualization, video demos 1 and 2 are played at 0.1x speed.

Demos 3 and 4 showcase the rendered images, optical flow, and depth by our SplatFlow in dynamic driving scenarios from the Waymo and KITTI datasets. 
In these videos, the first row shows the G.T RGB images.
The second row displays the rendered RGB images, the third row shows the rendered optical flow, and the fourth row presents the rendered depth, all generated by our SplatFlow.
For a clearer visualization, video demos 3 and 4 are played at 0.1x speed.

Demo 5 showcases the 3D motion prediction of LiDAR points within the NMFF.
The color wheel in the top right corner represents the flow magnitude through color intensity and flow direction via the angle.
For clearer visualization, video Demo 5 is played at 0.1x speed.
%

{
    \small
    \bibliographystyle{ieeenat_fullname}
    \bibliography{main}
}
\end{document}